\documentclass[journal]{IEEEtran}
\usepackage[utf8]{inputenc}
\DeclareUnicodeCharacter{2060}{\nobreak}
\usepackage{hyperref}       
\usepackage{url}            
\usepackage{booktabs}       
\usepackage{amsmath,amsfonts,amssymb}
\usepackage{algorithm}
\usepackage{algpseudocode}
\usepackage{array}
\usepackage[caption=false,font=normalsize,labelfont=sf,textfont=sf]{subfig}
\usepackage{textcomp}
\usepackage{stfloats}
\usepackage{url}
\usepackage{verbatim}
\usepackage{graphicx}
\usepackage{cite}
\usepackage{bm}
\usepackage{multirow}
\usepackage{amsmath}
\usepackage{float}
\usepackage[normalem]{ulem}
\useunder{\uline}{\ul}{}
\usepackage{xurl}
\usepackage{xcolor}
\usepackage{bbm}

\usepackage{xspace}
\makeatletter
\DeclareRobustCommand\onedot{\futurelet\@let@token\@onedot}
\def\@onedot{\ifx\@let@token.\else.\null\fi\xspace}
\def\eg{\emph{e.g}\onedot}

\newcommand{\Rmnum}[1]{\expandafter\@slowromancap\romannumeral #1@}
\makeatother

\begin{document}

\title{FontGuard: A Robust Font Watermarking Approach Leveraging Deep Font Knowledge}

\author{
Kahim~Wong,~Jicheng~Zhou,~Kemou~Li,~Yain-Whar~Si, \\Xiaowei~Wu,~and~Jiantao~Zhou,~\IEEEmembership{Senior~Member,~IEEE
\thanks{
This work was supported in part by Macau Science and Technology Development Fund under SKLIOTSC-2021-2023, 0022/2022/A1, and 0014/2022/AFJ; in part by Research Committee at University of Macau under MYRG-GRG2023-00058-FST-UMDF and MYRG2022-00152-FST; in part by the Guangdong Basic and Applied Basic Research Foundation under Grant 2024A1515012536. K. Wong, J. Zhou, K. Li, X. Wu, and J. Zhou are with the State Key Laboratory of Internet of Things for Smart City, and also with the Department of Computer and Information Science, Faculty of Science and Technology, University of Macau, China. Y.-W. Si is with the Department of Computer and Information Science, Faculty of Science and Technology, University of Macau, China. Emails: \{yc37437, mc35093, yc47912, fstasp, xiaoweiwu, jtzhou\}@um.edu.mo. \emph{(Corresponding author: Jiantao Zhou.)}
}
}
}

\markboth{IEEE Transactions on Multimedia}%
{Shell \MakeLowercase{\textit{et al.}}: A Sample Article Using IEEEtran.cls for IEEE Journals}


\maketitle

\begin{abstract}
The proliferation of AI-generated content brings significant concerns on the forensic and security issues such as source tracing, copyright protection, etc, highlighting the need for effective watermarking technologies. Font-based text watermarking has emerged as an effective solution to embed information, which could ensure copyright, traceability, and compliance of the generated text content. Existing font watermarking methods usually neglect essential font knowledge, which leads to watermarked fonts of low quality and limited embedding capacity. These methods are also vulnerable to real-world distortions, low-resolution fonts, and inaccurate character segmentation. In this paper, we introduce FontGuard, a novel font watermarking model that harnesses the capabilities of font models and language-guided contrastive learning. Unlike previous methods that focus solely on the pixel-level alteration, FontGuard modifies fonts by altering hidden style features, resulting in better font quality upon watermark embedding. We also leverage the font manifold to increase the embedding capacity of our proposed method by generating substantial font variants closely resembling the original font. Furthermore, in the decoder, we employ an image-text contrastive learning to reconstruct the embedded bits, which can achieve desirable robustness against various real-world transmission distortions. FontGuard outperforms state-of-the-art methods by +5.4\%, +7.4\%, and +5.8\% in decoding accuracy under synthetic, cross-media, and online social network distortions, respectively, while improving the visual quality by 52.7\% in terms of LPIPS. Moreover, FontGuard uniquely allows the generation of watermarked fonts for unseen fonts without re-training the network. The code and dataset are available at \href{https://github.com/KAHIMWONG/FontGuard}{https://github.com/KAHIMWONG/FontGuard}.                

  
 

 
\end{abstract}


\section{Introduction}

\IEEEPARstart{R}{egulations} on the watermarking of AI-generated content have emerged due to the rise of large language models and concerns over potential copyright risks \cite{chregular, barrett2023identifying, usai, song2025protecting, huang2025geometrysticker, luo2023copyrnerf}. Advocates stress the significance of transparency, traceability, and compliance through the implementation of watermarking techniques to identify such content as AI-produced. Additionally, integrating extra data, \eg, hyperlinks, metadata, or identifiers into text via watermarking provides practical and commercial benefits \cite{xiao2018fontcode, qi2019robust, begum2020digital, yang2023language, yang2023autostegafont, huang2024robust, yao2024embedding}. Text watermarking not only enhances access to information but also protects copyright, authenticates content, and maintains the integrity of sensitive materials, thereby fostering a more secure and accountable digital environment.

\begin{figure*}[tb]
\centering
\includegraphics[width=0.9\textwidth]{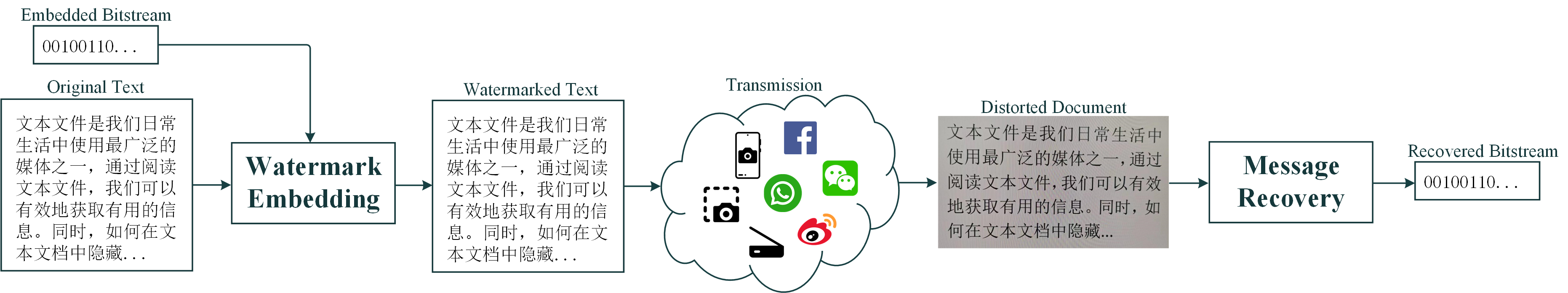}
\caption{An overview of text watermarking.}
\label{fig:watermark_process_general}
\end{figure*}

The watermark embedding and message recovery process for text watermarking is depicted in Fig.~\ref{fig:watermark_process_general}. The process involves embedding a bitstream into a document, which can then be distributed through various channels, potentially subject to distortions \cite{yang2023language, yang2023autostegafont, sun2016processing, sun2018robust, sun2020robust, sun2021optimal, wu2023robust, liu2023generating}. The hidden message is subsequently decoded from the distorted document. Numerous schemes have been developed for text watermarking \cite{brassil1995electronic, kim2003text, alattar2004watermarking, atallah2002natural, topkara2006words, topkara2006hiding, yang2020vae, yoo2023robust, peng2023text, zhu2018hidden, tancik2020stegastamp, luo2020distortion, zhong2020automated, fang2022end, qin2023print, 9956019, ge2023robust, ge2023screen}. Text watermarking schemes can be broadly categorized into four classes \cite{yang2023language}: format-based \cite{brassil1995electronic, kim2003text, alattar2004watermarking}, linguistic-based \cite{atallah2002natural, topkara2006words, topkara2006hiding, yang2020vae, yoo2023robust, peng2023text}, image-based \cite{zhu2018hidden, tancik2020stegastamp, luo2020distortion, zhong2020automated, fang2022end, qin2023print, 9956019, ge2023robust, ge2023screen, zhang2019steganogan, jing2021hinet, jia2021mbrs}, and font-based \cite{xiao2018fontcode, qi2019robust, yang2023language, yang2023autostegafont, huang2024robust, yao2024embedding} methods. Format-based watermarking methods focus on modifying the format of electronic documents \cite{brassil1995electronic, kim2003text, alattar2004watermarking}. These methods embed information by adjusting the horizontal \cite{brassil1995electronic, kim2003text} or vertical \cite{alattar2004watermarking} spacing between characters. Nevertheless, their practical application in real-world scenarios is limited, since original electronic documents are required for decoding hidden bits. Linguistic-based algorithms modify text content for watermark embedding, such as substituting specific words \cite{topkara2006hiding} or sentences \cite{topkara2006words}, altering grammatical structures \cite{atallah2002natural}, or creating hidden patterns \cite{yang2020vae, peng2023text, yoo2023robust}. Linguistic-based methods are inherently robust to diverse distortions since the degradation in the RGB domain does not affect content in the linguistic domain, unless the characters cannot be recognized. However, this approach could lead to the sentence incoherence, making it unsuitable for situations that require absolute sentence integrity, \eg, commercial contracts, government documents, and academic papers. In image-based watermarking, documents are processed as images, applying pixel \cite{wu2004data, cu2020robust}, color \cite{borges2008robust, loc2018document}, and learnable perturbation \cite{cu2019hiding, ge2023robust, ge2023screen} to the document images. More recently, the joint optimization of the watermarking encoder and decoder within an end-to-end framework has demonstrated effectiveness \cite{zhu2018hidden, tancik2020stegastamp, luo2020distortion, zhong2020automated, fang2022end, qin2023print, 9956019, ge2023robust, ge2023screen, zhang2019steganogan, jing2021hinet}. However, the visual gaps between natural and document images often result in quality and accuracy degradation when image-based algorithms are applied to document images.  







In addition to the aforementioned schemes, font-based watermarking algorithms have emerged as a promising solution \cite{xiao2018fontcode, qi2019robust, yang2023language, yang2023autostegafont, huang2024robust, yao2024embedding}, better maintaining the integrity of sentences while having a larger embedding capacity. Despite the advancements achieved by existing font-based methods, there is still significant room for further improvement, especially in the way of generating and decoding watermarked fonts. Regarding perturbed font generation, current methods \cite{yang2023language, yang2023autostegafont, huang2024robust, yao2024embedding} focus solely on pixel-level alteration and overlook the global consistency of fonts, resulting in low-quality watermarked fonts and limited embedding capacity. For instance, the reported bit capacity per character (BPC) in \cite{yang2023language} and \cite{yang2023autostegafont} are only 0.5 and 1, respectively. Additionally, the above approaches suffer severe degradation of the decoding accuracy when faced with transmission noise across various media (\eg, print-camera shooting and screen-camera shooting) \cite{yang2023autostegafont}, and online social networks (OSNs) \cite{wu2022robust}, as well as challenges such as low-resolution fonts, complex backgrounds, and inaccurate character segmentation.

To tackle the aforementioned challenges, in this paper, we introduce FontGuard, a robust and imperceptible text watermarking solution with high embedding capacity. First of all, to create high-quality watermarked fonts, we propose to synthesize fonts using a font manifold derived from a well-trained deep font model. Different from existing works \cite{yang2023language, yang2023autostegafont}, which solely focus on pixel-level alteration, our FontGuard modifies fonts by adjusting hidden style features. This strategy ensures the global consistency in altering character appearances, thereby enhancing the quality of the generated fonts. Additionally, leveraging the font manifold allows the model to generate substantial font variants closely resembling the original font, which essentially increases our embedding capacity. We further propose FontGuard-GEN which exploits the font model's ability to generate new fonts, enabling the generation of watermarked fonts for unseen fonts without re-training the network. For the watermarking recovery process, we adopt a contrastive language-image pertaining (CLIP) training paradigm \cite{radford2021learning}, which utilizes language-guided contrastive learning to train the decoder network, instead of deploying a cross-entropy classifier. Such a paradigm, well-known for enhancing neural network representation abilities \cite{radford2021learning, zheng2023exif, wu2023generalizable}, enables the decoder to extract highly discriminative features resilient against real-world transmission distortions, thereby boosting the decoding accuracy. Ultimately, the FontGuard networks are trained in an end-to-end fashion to optimize both the visual quality and decoding accuracy of watermarked fonts. In summary, our contributions are threefold:              



\begin{itemize}
\item We present FontGuard, integrating deep font knowledge into an end-to-end watermarking framework. This approach yields high-quality watermarked fonts while substantially enhancing bit capacity through the generation of diverse font variants.

\item Our method employs the CLIP paradigm for decoding hidden bits from perturbed fonts, exhibiting remarkable resilience against real-world transmission distortions.

\item Extensive experiments validate the robustness of FontGuard against synthetic, cross-media, and OSNs distortions, surpassing state-of-the-art methods by +5.4\%, +7.4\%, and +5.8\% while improving the visual quality by 52.7\%. Furthermore, our method increases the BPC by \( 4 \times \) and demonstrates exceptional generalizability by effectively generating watermarked fonts for a wide range of unseen fonts in a training-free manner.

\end{itemize}

The remaining content of this paper is organized as follows. Sec.~\ref{sec:related} reviews the relevant works in font watermarking. Sec.~\ref{sec:method} gives the details of the proposed FontGuard. Sec.~\ref{sec:experiment} presents extensive experimental results, validating FontGuard's superior performance. Finally, Sec.~\ref{sec:conclusion} concludes.      


\section{Related Works}\label{sec:related}

In this section, we first provide a list of frequently-used acronyms with their corresponding full forms in Table~\ref{table:acronyms}, then have a brief review of related works on the font watermarking and the font modeling.


\begin{table}[ht]
\vspace{-10pt}
\centering
\caption{Table of Acronyms}
\renewcommand{\arraystretch}{1.2}
\begin{tabular}{|>{\raggedright\arraybackslash}p{2cm}|>{\raggedright\arraybackslash}p{6cm}|}
\hline
\textbf{Acronym} & \textbf{Full Form} \\
\hline
OSNs & Online Social Networks \\
\hline
RGB & Red Green Blue \\
\hline
BPC & Bit Capacity per Character \\
\hline
CLIP & Contrastive Language-Image Pretraining \\
\hline
OCR & Optical Character Recognition \\
\hline
SVG & Scalable Vector Graphics \\
\hline
MLP & Multilayer Perceptron \\
\hline
SOTA & State-of-the-art \\
\hline
\end{tabular}
\label{table:acronyms}
\vspace{-10pt}
\end{table}

\begin{figure*}[tb]
\centering
\includegraphics[width=0.88\textwidth]{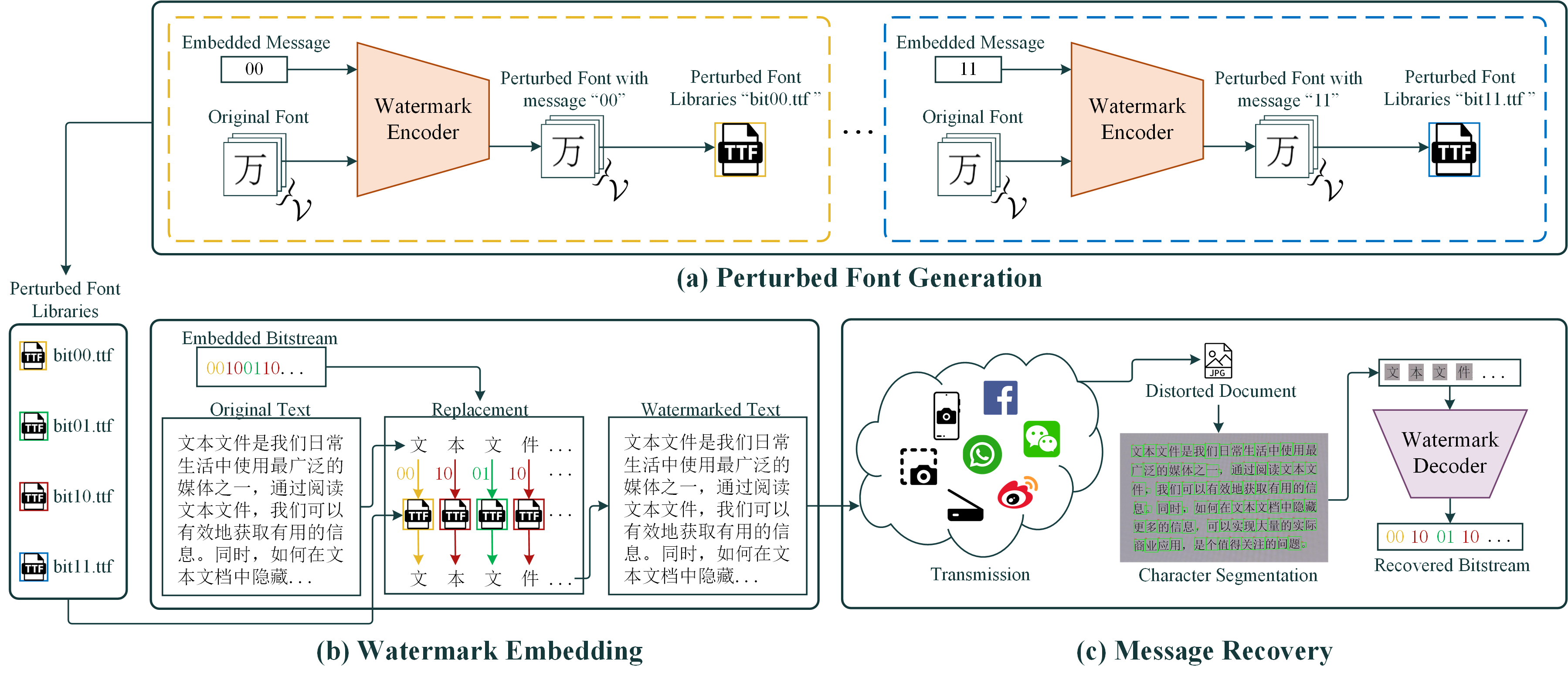}
\caption{System model for our font-based text watermarking.}
\label{fig:watermark_process}
\end{figure*}

\subsection{Font Watermarking}

Font-based watermarking scheme has been one of the most promising solutions for text watermarking, which embeds information in documents by perturbed font replacement. Along this line, FontCode \cite{xiao2018fontcode} leverages the font manifold \cite{campbell2014learning} and introduces noises in the feature domain to create perturbed fonts. Each character is then assigned a classifier for the message recovery. RATD \cite{qi2019robust} designs watermarked font libraries manually by altering stroke positions and uses a template-matching algorithm for decoding the embedded message. However, both FontCode and RATD heavily rely on accurate optical character recognition (OCR) algorithms for the character recognition and classifier assignment, or matching templates to recover the hidden message. EMCC \cite{yao2024embedding} manually creates a perturbed font and trains a font model to transfer these perturbations to other fonts, while using a font variant match algorithm to recover the message. LUFW \cite{yang2023language} and ASM \cite{huang2024robust} embed messages by modifying character edges to alter centroid positions and skeleton masses. AutoStegaFont \cite{yang2023autostegafont} implements a two-stage approach that begins with a joint training of the encoder and decoder, followed by the optimization of scalable vector graphics (SVG) commands under the supervision of both networks. Nevertheless, AutoStegaFont faces challenges in generalizing to unseen fonts, having to go through the tedious re-training for new fonts. Moreover, as will become clearer later, these methods are vulnerable to real-world distortions, resulting in inferior decoding accuracy. 


\subsection{Font Modeling}
Font modeling plays a crucial role in the font-based watermarking, and there are two modalities to represent the appearance of fonts: image and SVG.  Image-based font models \cite{park2021few, xie2021dg, wang2023cf, yao2023vq, 10356848} receive and generate fonts using pixel values that indicate the intensity of each color in the image. On the other hand, SVG-based models \cite{campbell2014learning, carlier2020deepsvg, wang2021deepvecfont, wang2023deepvecfont} utilize a set of commands that specify the coordinates of control points for straight lines or bezier curves, which form the boundary of fonts. When handling characters with more intricate structures, \eg, Chinese or Korean characters, image-based glyph models exhibit superiority in terms of the quality of the generated fonts. This advantage stems from the fact that the input complexity remains consistent regardless of the character structures. In our FontGuard, we adopt the image-based font model, due to its superiority when handling complex characters.     

\section{Method}\label{sec:method}

In this section, we elaborate on the details of FontGuard, our solution for robust, high-capacity, and imperceptible font watermarking. We begin by explaining the motivation, followed by the system model of our font watermarking scheme. We then give the details on the FontGuard network architecture and present the model objective and training strategies.      

\subsection{Motivation}
\label{sec: motivation}
Font models, as demonstrated in prior studies \cite{park2021few, xie2021dg, wang2023cf, yao2023vq, 10356848}, excel in generating high-quality fonts. These models can create imperceptible variants from an original font by perturbing the hidden features. In contrast to creating watermarked fonts through pixel-level alterations \cite{yang2023language, yang2023autostegafont}, modifying fonts by adjusting hidden features could better ensure the global consistency of appearances and enhance the font quality. By perturbing the hidden feature in different ways, we can generate many font variants, significantly improving the watermark embedding capacity. Also,  exploiting the capability of the font model to create new fonts facilitates the generation of watermarked fonts for unseen ones. Regarding the decoder, the CLIP paradigm has shown promise in learning robust features for classification tasks \cite{radford2021learning, wu2023generalizable}. Given that watermarking decoding can also be formulated as a classification problem, we adopt the language-guided contrastive learning for potentially improving the decoding performance and generalizability to various distortions. 

\subsection{System Model}
The embedding and message recovery process of our proposed font-based watermark scheme is illustrated in Fig.~\ref{fig:watermark_process}. Since the watermark embedding is achieved by replacing the original font with its perturbed version, we depict the way of generating the perturbed font in Fig.~\ref{fig:watermark_process} (a). The original font, encompassing all characters in the vocabulary \(\mathcal{V}\), along with a specified embedded message (\eg, ``00"), is input into a watermark encoder. This encoder generates a set of perturbed characters that look similar to the original font, but carry the hidden message ``00". The perturbed characters are then encapsulated into a font library for the future use. The above process is repeated for all hidden messages, and a set of perturbed font libraries can be obtained. To embed the watermark into a document, bits from an encoded bitstream are assigned to characters in the document based on the BPC, as shown in Fig.~\ref{fig:watermark_process} (b). For instance, if BPC is 2, then every 2 bits will be embedded into one character in a sequential order, by replacing those from the specified perturbed font libraries. The watermarked document can then be distributed through various channels, with a certain level of distortions \cite{yang2023language, yang2023autostegafont, sun2016processing, sun2018robust, sun2020robust, sun2021optimal, wu2023robust, liu2023generating}. To recover the hidden message, characters are segmented from the distorted document, and the hidden bits are decoded by the watermark decoder, as demonstrated in Fig.~\ref{fig:watermark_process} (c).      

We are now ready to present the details of the watermark encoder and decoder for our FontGuard.

\begin{figure*}
\centering
\includegraphics[width=0.95\textwidth]{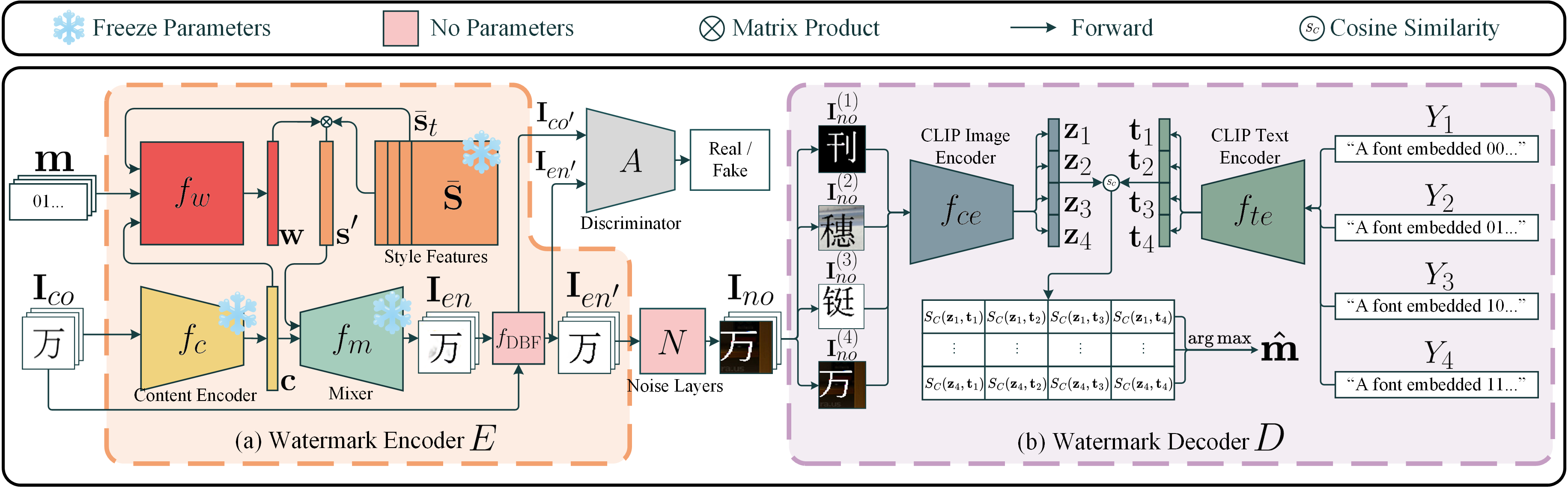}
\caption{The watermark encoder/decoder in FontGuard. The model employs an end-to-end training approach, where the watermarking encoder and decoder are jointly trained alongside noise layers. The encoder consists of two main components: a well-trained font model and a learnable weight module. The font model provides a manifold for perturbed fonts, while the weight module encourages the encoder to generate appropriate fonts by learning. The decoder, based on a pre-trained CLIP model, is designed to maximize the similarity between a noisy, watermarked font image and its corresponding textual label.}
\label{fig:model_overview}
\end{figure*}

\subsection{Watermark Encoder/Decoder in FontGuard} 
\label{subsec: model overview}

The architecture of the watermark encoder \(E\) and watermark decoder \(D\) in our proposed FontGuard is illustrated in Fig.~\ref{fig:model_overview}, where we also include the noise layers \(N\) and an adversarial discriminator \(A\). Specifically, the encoder \(E\) receives a cover font image \(\mathbf{I}_{co}\) and a secret message \(\mathbf{m}\), and attempts to generate a perturbed version of the font \(\left.\mathbf{I}_{en}=E(\mathbf{I}_{co}, \mathbf{m})\right.\). The noise layers \(N\) augment the perturbed font by simulating the transmission noises and produce \(\mathbf{I}_{no} = N(\mathbf{I}_{en})\), which is then fed into the \(D\) to recover the embedded message \(\left.\hat{\mathbf{m}}=D(\mathbf{I}_{no})\right.\). Concurrently, the discriminator \(A\) differentiates between \(\mathbf{I}_{co}\) and \(\mathbf{I}_{en}\), ensuring the perturbations remain subtle. The entire network is jointly trained, of which \(E\) is updated through backpropagation from \(D\).


\subsubsection{Encoder \(E\)}
\label{subsec: enc}

We now delve into the details of the watermark encoder \(E\), as depicted in Fig.~\ref{fig:model_overview} (a). Since the encoder is integrated with a well-trained font model, we first introduce the structure of the font model, and formulate the font manifold derived from it. We then explain how we use the font manifold to generate perturbed fonts in a learnable fashion. Finally, we present the solution to mitigate the subtle noise in the generated fonts.

\begin{figure}
\centering
\includegraphics[width=6cm]{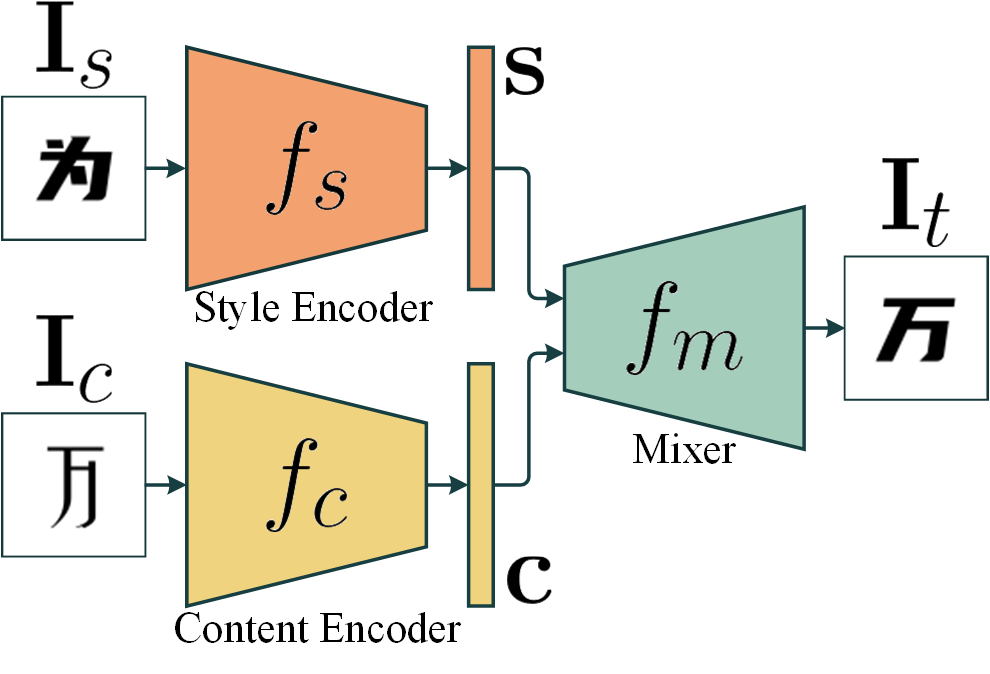}
\caption{The style-content disentanglement structure of font models}
\label{fig:font_model}
\end{figure}

The prevailing architecture of font models is illustrated in Fig.~\ref{fig:font_model}, which adopts a structure that decouples style and content features\cite{park2021few, xie2021dg, wang2023cf, yao2023vq, 10356848}. The objective of a font model is to craft a character in a specified font style. This is achieved by using two inputs: a content image \( 
\mathbf{I}_{c}\), which is the target character in any font, and a style image \(\mathbf{I}_{s}\), showing a random character in the desired font style. The model employs two encoders: the content encoder \(f_c\) extracts content features \(\mathbf{c}\) from \(\mathbf{I}_{c}\), and the style encoder \(f_s\) derives style features \(\mathbf{s}\) from \(\mathbf{I}_{s}\). These features are then combined in a mixer \(f_m\) to generate the character in the desired style:  
\begin{equation}
    \mathbf{I}_t=f_m(f_c(\mathbf{I}_c), f_s(\mathbf{I}_s)).
    \label{eq:font model}
\end{equation}

Given a well-trained font model characterized by a style-content decoupled structure, we can demonstrate the presence of a font manifold within the model. This font manifold enables the generation of different font styles for a given character. When a content image of the character is provided, the font manifold can be formulated by treating the style feature as a variable while keeping the other model components fixed. For a given content image \( \mathbf{I}_c \), and a style feature \(\mathbf{s} \in \mathbb{R}^d\), the font manifold \(\mathcal{F}\) can be expressed as:
\begin{equation}\label{font manifold}
    \mathcal{F} = \Big\{f_m(f_c(\mathbf{I}_c), \mathbf{s}) \mid \mathbf{s} \in \mathbb{R}^d\Big\}.
\end{equation}

To generate characters in a specific font style \( \mathbf{I}_{s^{\prime}} \), a specific style feature \(\mathbf{s}^{\prime}\) is input:
\begin{equation}\label{font image}
    \mathbf{I}_{s^{\prime}} = f_m(f_c(\mathbf{I}_c), \mathbf{s}^{\prime}).
\end{equation}


With the concept of the font manifold in mind, we now explain how to integrate the font model into the watermark encoder. Specifically, the font model is integrated into the watermark encoder \(E\), as shown in  Fig.~\ref{fig:model_overview} (a), where networks are colored identically as in Fig.~\ref{fig:font_model} for clarity. An extra lightweight network \(f_w\) in red is included to learn a proper style feature \(\mathbf{s}^{\prime}\) for generating watermarked characters. In practice, the font model may not always generalize effectively, potentially failing to map arbitrary style features to high-quality font images. It is empirically known that the style feature can be approximated by those present in the training dataset \cite{park2021few}. To enforce this proximity, the new style features \(\mathbf{s}^{\prime}\) are derived by blending existing font styles with the weight \(\mathbf{w}\) and the network \(f_w\) is responsible to learn the proper \(\mathbf{w}\). Concretely, the mean style feature \(\bar{\mathbf{s}}_i\) of the \(i\)-th existing font is computed by averaging the style features across all characters in the vocabulary:
\begin{equation}
    \bar{\mathbf{s}}_i = \frac{1}{|\mathcal{V}|}\sum_{v=0}^{|\mathcal{V}|}{\mathbf{s}_{i,v}},
    \label{font manifold}
\end{equation}
where \( |\mathcal{V}| \) represents the total number of characters and \( \mathbf{s}_{i,v} \) denotes the style feature of the \(v\)-th character in the \(i\)-th font. For all the existing fonts, along with the target font to watermark indexing with \(t\), we have matrix \(\bar{\mathbf{S}}\) composed of features \( \left.\bar{\mathbf{S}} = [\bar{\mathbf{s}}_1, \bar{\mathbf{s}}_2, \dots,  \bar{\mathbf{s}}_n, \bar{\mathbf{s}}_t]\in\mathbb{R}^{d\times (n+1)}\right.\). Subsequently, a new style feature  \( \mathbf{s}^{\prime} = \bar{\mathbf{S}}\mathbf{w} \) is formulated through a convex combination of mean style features, where \( \left.\mathbf{w} = [w_1, w_2, \dots, w_n, w_{t}]^{\top}\in\mathbb{R}^{n+1}\right. \) is the weight vector subject to \( \left. \sum_i w_i=1 \right.\). Since the mean style features \(\bar{\mathbf{S}}\) can be pre-computed, \(f_s\) is omitted, retaining \(f_c\) and \(f_m\). To facilitate the learning process, a lightweight network \(f_w\) is introduced to learn the blending weights \(\mathbf{w}\), with \( \bar{\mathbf{s}}_t \), \( \mathbf{c} \), and \( \mathbf{m} \) as inputs. These features are first mapped to a uniform dimension \( \mathbb{R}^{d} \), normalized, and aggregated before being mapped to \(\mathbf{w}\) by a multilayer perceptron (MLP) network. The resulting \( \mathbf{w} \) is determined as:
\begin{equation}
    \mathbf{w} =\mathrm{softmax}(f_w(\bar{\mathbf{s}}_t, \mathbf{c}, \mathbf{m})/\tau_{w}),
    \label{eq:learnable w}
\end{equation}
where \(\tau_{w}\) is the temperature and the softmax function is used to ensure the sum of weights is equal to 1. The perturbed font image \( \mathbf{I}_{en} \) is then synthesized as
\begin{equation}
     \mathbf{I}_{en}=f_m(f_c(\mathbf{I}_{co}), \bar{\mathbf{S}}\mathbf{w}).
    \label{eq:enc img}
\end{equation} Also, within this structure, while \(f_c\) and \(f_m\) remain frozen, only the parameters of \(f_w\) are updated.


\begin{figure}[H]
\centering
\includegraphics[width=6cm]{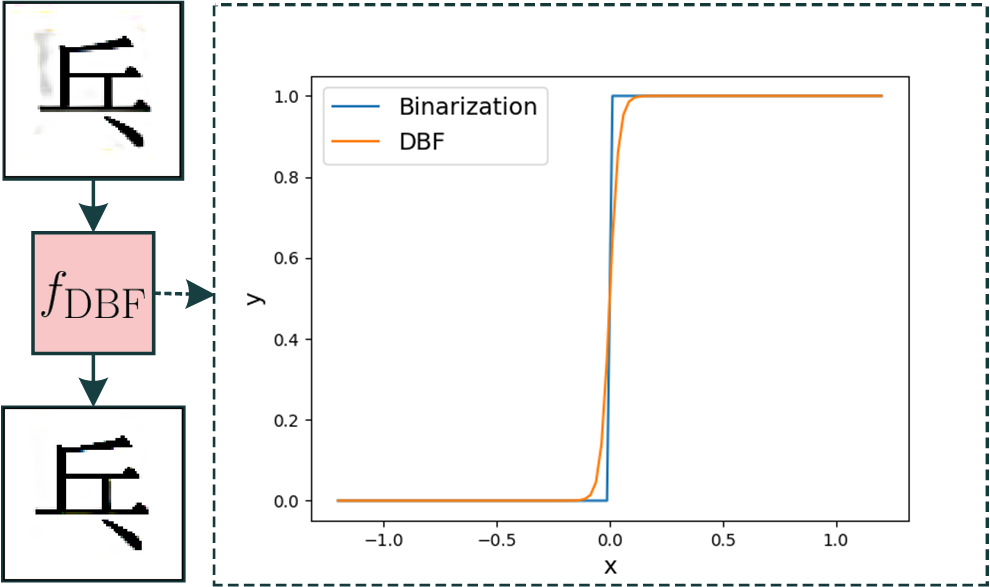}
\caption{The behavior of Differential Binarization Filter}
\label{fig:dbf}
\end{figure}

Upon constructing the font model-based watermark encoder \(E\), we observe that the font model inherently introduces background noise when generating fonts via style feature interpolation, as illustrated by the upper fonts in Fig.~\ref{fig:dbf}. More seriously, the decoder network could prioritize the subtle noise over character-level appearance when retrieving embedded bits, resulting in trivial solutions.
To resolve this problem, we notice that the pixel intensity of noise is weaker than the pixel on the character. Applying standard binarization to \(\mathbf{I}_{en}\) is a potential solution; but it is not differentiable. To address this, we use a scaled sigmoid function as an approximation which acts as a pixel-wise filter to refine the perturbed image by removing artifacts. This module is termed the differentiable binarization filter (DBF), mathematically expressed as:
\begin{equation}
    \mathbf{I}_{en^{\prime}}=f_\mathrm{DBF}(\mathbf{I}_{en})=\frac{1}{1+\exp(-k(\mathbf{I}_{en}-\sigma))},
    \label{eq:db}
\end{equation} in which \(k\) adjusts the sharpness of binarization and \(\sigma\) denotes the binarization threshold. As illustrated in Fig.~\ref{fig:dbf}, the behavior of DBF closely mirrors the standard binarization, with the added advantage of being differentiable. Given that pixel value differences exist before and after applying DBF, both \(\mathbf{I}_{en}\) and \(\mathbf{I}_{co}\) undergo DBF before feeding into the discriminator \(A\). This step ensures \(A\) does not resort to trivial solutions in classifying \(\mathbf{I}_{co}\) and \(\mathbf{I}_{en}\).




\subsubsection{Noise Layers \(N\)}
\label{subsec: noise}


Between the encoder \(E\) and decoder \(D\), we incorporate a series of differentiable noise layers \(N\). These layers are designed to enhance the decoding robustness by simulating the noises encountered in real-world scenarios. Following \cite{tancik2020stegastamp}, our noise layers address both pixel-level and spatial-level distortions, including blurring, cropping, resizing, noise addition, perspective transformation, and color manipulation. During each training step, three random augmentations are selected to form an augmentation chain:  
\begin{equation}
    \mathbf{I}_{no} = N(\mathbf{I}_{en^{\prime}}) = (N_2 \circ N_1 \circ N_0)(\mathbf{I}_{en^{\prime}}),
    \label{eq:db}
\end{equation} where \(N_0\), \(N_1\), and \(N_2\) are sampled augmentations. In addition to these augmentations, we introduce another specific augmentation to align with common document transmission scenarios. Documents, such as certificates, often feature patterns and backgrounds for visual appeal. While previous works primarily focus on plain-background documents \cite{qi2019robust, yang2023language, yang2023autostegafont}, our approach integrates complex backgrounds into the watermarked fonts, making our decoder background-agnostic. To achieve this, since the watermarked font \(\mathbf{I}_{en^{\prime}}\) is on a pristine white background, we use standard binarization to extract masks for the font \(\mathbf{M}_{f}\) and the background \(\mathbf{M}_{b}\). A random natural image \(\mathbf{I}_{b}\) is then superimposed onto \(\mathbf{I}_{en^{\prime}}\), which is formulated as:
\begin{equation}
    N_\mathrm{BGA}(\mathbf{I}_{en^{\prime}}) = \mathbf{I}_{en^{\prime}}\odot\mathbf{M}_{f} + \mathbf{I}_{b}\odot\mathbf{M}_{b},
    \label{eq:db}
\end{equation} where \(\odot\) denotes the element-wise product. Note that the background augmentation is included before the other three sampled augmentations.



\subsubsection{Decoder \(D\)}
\label{subsec: dec}
A decoding network is employed mainly to recover the hidden message in watermarked fonts, utilizing a CLIP structure to further enhance the decoding robustness against diverse post-processing and transmission noises. As can be seen from Fig.~\ref{fig:model_overview} (b), a set of textual labels \( \{Y_{1},\dots, Y_{a^2}\} \) is indexed by \( \mathcal{Y} = \{1,\dots, a^2\}\), which \(a\) denotes the BPC. The textual label \(Y\) is designed in a form ``a font embedded \(\mathbf{m}\)'', where \(\mathbf{m}\) denotes the associated message. In the training phase, \(\mathbf{m}\) is sampled from a uniform distribution, i.e., \(\mathbf{m} \sim \text{Uniform}(\{0,1\}^a)\). Both \(\mathbf{I}_{no}\) and the corresponding textual label \(Y\)  are fed into encoders \(f_{ce}\) and \(f_{te}\) to extract visual and textual features \(\mathbf{z}=f_{ce}(\mathbf{I}_{no})\) and \(\mathbf{t}=f_{te}(Y)\). A contrastive loss is used to maximize the similarity between paired image and textual feature and minimize the one between the unmatched pairs. In the inference phase, a well-trained CLIP image encoder \(f^{*}_{ce}\) is used to infer the embedded message in a font image \(\mathbf{I}_{no}\). Since the textual labels are fixed, \(f_{te}\) is discarded, and only \(f^{*}_{ce}\) and textual features \( \mathcal{T} = \{\mathbf{t}_1, \mathbf{t}_2, \dots, \mathbf{t}_{a^2} \}\) are required. More specifically, to infer message \(\hat{\mathbf{m}}\), the image feature of a font image is extracted by \(\mathbf{z} = f^{*}_{ce}(\mathbf{I})\), and then cosine similarity is computed between the image feature and all the textual features in \(\mathcal{T}\):  
\begin{equation}
    S_C(\mathbf{z}, \mathbf{t}_k) = \frac{\langle \mathbf{z} , \mathbf{t}_k\rangle}{\|\mathbf{z}\|_2 \|\mathbf{t}_k\|_2}.
    \label{eq:enc img}
\end{equation} Then, the message \(\hat{\mathbf{m}}\) can be recovered by
\begin{equation}
    \hat{\mathbf{m}} = \underset{\mathbf{t}_k \in \mathcal{T}}{\arg\max}\, S_C(\mathbf{z}, \mathbf{t}_k).
    \label{eq:enc img}
\end{equation}


\begin{figure}[H]
\centering
\includegraphics[width=8cm]{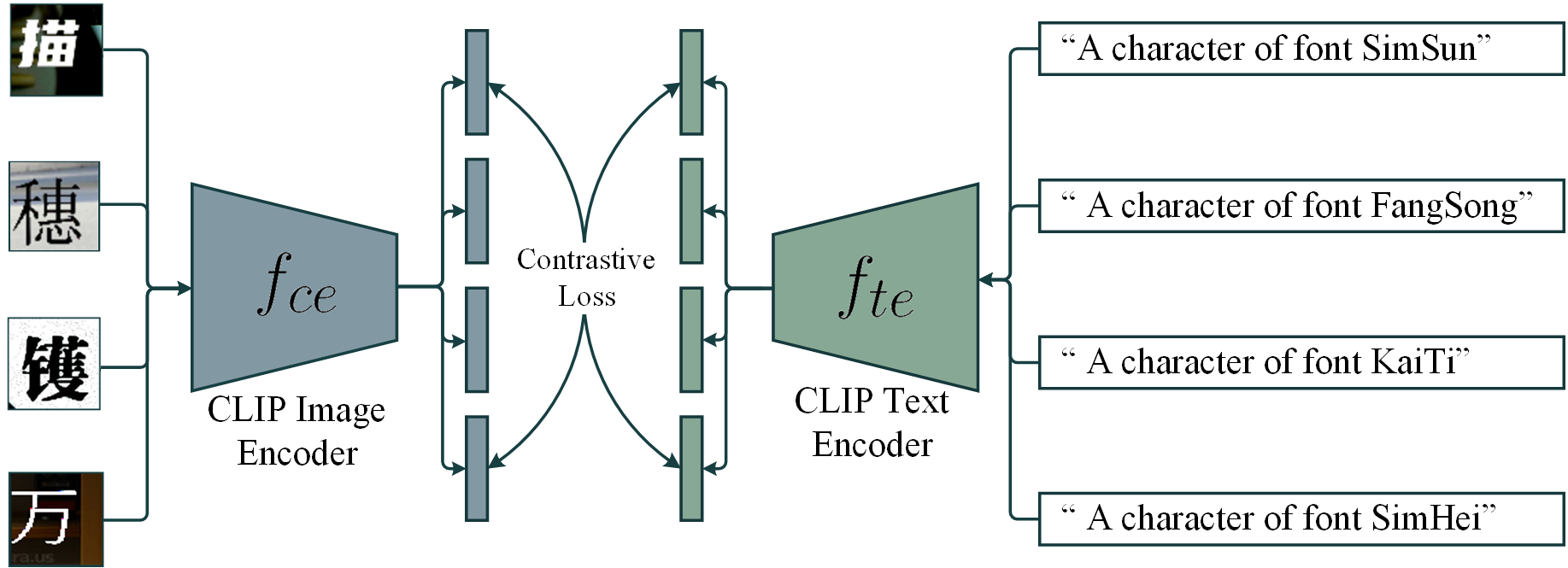}
\caption{An example of the font classification task. }
\label{fig:pretrain_clip}
\end{figure}

In the above process of incorporating CLIP, it should be noted that simply deploying a CLIP decoder into the end-to-end model fails to converge. The reason is found to be the decoder fails to distinguish fonts with different bits embedded in the initial stage of training. To overcome this issue, our watermarking decoder is initialized on a font classification task, which is to classify augmented font images into correct font labels. Fig.~\ref{fig:pretrain_clip} shows an example of the font classification task for the pre-training of CLIP. With such a pre-training, the network has a better starting point and hence convergence can be readily achieved.    


Upon the introduction of the watermark Encoder/Decoder in FontGuard, we now present FontGuard-GEN, a generalized model building on the FontGuard architecture, which generates watermarked fonts for unseen fonts without re-training. While the original FontGuard network is trained on a specific font, FontGuard-GEN is trained on a wide range of fonts for generalization. Because of the extension of the training fonts, we observe that the watermark decoder \(D\) is not able to extract embedded bits across different watermarked fonts.  Instead, we introduce the style feature \(\bar{\mathbf{s}}_t\) of fonts, which provides information about its attributes, such as serif-ness and thickness, as a prompt to the decoder. This style prompt allows \(D\) to conditionally extract features for the bits recovery based on the specific attributes of the font. Formally, the visual feature \(\mathbf{z}\) extracted by \(D\) is combined with \(\bar{\mathbf{s}}_t\) using a MLP \(f_{sp}\) that maps \(\bar{\mathbf{s}}_t\) to match the dimensionality of \(\mathbf{z}\). The modified feature \(\mathbf{z}^{\prime}\) is then computed as:
\begin{equation}
    \mathbf{z}^{\prime} = \mathbf{z} + f_{sp}(\bar{\mathbf{s}}_t).
    \label{eq:enc img}
\end{equation} With this modification, only the original visual feature \(\mathbf{z}\) is replaced with \(\mathbf{z}^{\prime}\), while keeping the other FontGuard networks unchanged, FontGuard-GEN enables accurate bit recovery from perturbed fonts across various fonts.


\subsection{Training Strategy and Objectives}
\label{subsec: obj}

Four losses are adopted to train FontGuard. Image reconstruction loss \(\mathcal{L}_{IR}\) and adversarial loss \(\mathcal{L}_{G}\) are used to ensure the quality of fonts. Style disparity loss \(\mathcal{L}_{SD}\) is employed to prevent the non-convergence problem at the initial stage of training by forcing the encoder to generate distinctive fonts according to different messages. Also, message reconstruction loss \(\mathcal{L}_{MR}\) is used to ensure decoding accuracy. For the generalized model FontGuard-GEN, style consistent loss \(\mathcal{L}_{SC}\) is introduced to overcome the style feature inconsistency when various fonts are trained. 

The training of FontGuard is summarized in Algorithm \ref{alg:training}. More specifically, we divide the training into two stages. In Stage \Rmnum{1}, \(\mathcal{L}_{MR}\) and \(\mathcal{L}_{SD}\) are applied, and the encoder is encouraged to generate distinctive font images. The objective in Stage \Rmnum{1} is:
\begin{equation}
    \mathcal{L}_1 = \mathcal{L}_{MR}+\lambda_{SD}\mathcal{L}_{SD}.
    \label{stage 1 loss}
\end{equation}
After \(T_1\) training epochs, the training stage is shifted to Stage \Rmnum{2}, and losses \(\mathcal{L}_{MR}\), \(\mathcal{L}_{IR}\), and \(\mathcal{L}_{G}\) are applied. The overall objective in Stage \Rmnum{2} can be expressed as:
\begin{equation}
    \mathcal{L}_2 = \mathcal{L}_{MR} + \lambda_{IR}\mathcal{L}_{IR} + \lambda_{G}\mathcal{L}_{G}.
    \label{stage 2 loss}
\end{equation}

For the generalized model FontGuard-GEN, the term \( \mathcal{L}_{SC} \) is incorporated in both stages. Moreover, PCGrad \cite{yu2020gradient} is adopted to mitigate conflicts between \(\mathcal{L}_{IR}\) and \(\mathcal{L}_{MR}\) in the encoder, striking a better balance of these two objectives.

\begin{algorithm}
\caption{\sc{FontGuard Training}}
\begin{algorithmic}[1] 
    \State \textbf{Input:} Cover font image \(\mathbf{I}_{co}\), bit capacity \( a\),  total training epoch \(T\), Stage \Rmnum{1} training epoch \(T_1\), learning rate \(\eta\), initial networks $E,D,A$ with training variables \( \boldsymbol{\theta}_{E}, \boldsymbol{\theta}_{D}, \boldsymbol{\theta}_{A} \)

    \State \textbf{Output:} Trained networks \(E, D, A\).

    \For{\( i = 1 : T\)}
        \State \( \mathbf{m} \sim \mathrm{Uniform}(\{0, 1\}^{a})\)
        \State \( \mathbf{I}_{en} = E(\mathbf{I}_{co}, \mathbf{m}) \)
        \State \( \mathbf{I}_{no} = N(\mathbf{I}_{en}) \)
        \State \( \hat{\mathbf{m}} = D(\mathbf{I}_{no})\)

        \State Update \(\boldsymbol{\theta}_{D} \leftarrow \boldsymbol{\theta}_{D} + \eta \cdot \nabla\mathcal{L}_{MR}\)
        \State Update \(\boldsymbol{\theta}_{A} \leftarrow \boldsymbol{\theta}_{A} + \eta \cdot \nabla\mathcal{L}_A\)
        
        \If{\(i \leq T_1\)} \Comment{Stage \Rmnum{1}}
            \State Update \(\boldsymbol{\theta}_{E} \leftarrow \boldsymbol{\theta}_{E} + \eta \cdot \text{PCGrad}(\mathcal{L}_{SD}, \mathcal{L}_{MR})\)
        \Else \Comment{Stage \Rmnum{2}}
            \State Update \(\boldsymbol{\theta}_{E} \leftarrow \boldsymbol{\theta}_{E} + \eta \cdot \text{PCGrad}(\mathcal{L}_{IR}, \mathcal{L}_{G}, \mathcal{L}_{MR})\)
        \EndIf

    \EndFor

\end{algorithmic}
\label{alg:training}
\end{algorithm}

We now explain the details of the loss terms  \(\mathcal{L}_{SD}\), \(\mathcal{L}_{MR}\), \(\mathcal{L}_{SC}\), \(\mathcal{L}_{IR}\), and \(\mathcal{L}_{G}\). To this end, we first denote \( b \in \mathcal{B}=\{1,\dots, B\}\) as the index set of a mini-batch. A batch of cover fonts and embedded bits is represented as \( \{\mathbf{I}^{(b)}_{co}, \mathbf{m}_{b}\}^{B}_{b=1} \). Encoded fonts are also indexed by \(b\), such that \(\mathbf{I}^{(b)}_{en^{\prime}}=E({\mathbf{I}^{(b)}_{co}}, \mathbf{m}_{b}) \) and \( \mathbf{I}^{(b)}_{no}=N(\mathbf{I}^{(b)}_{en^{\prime}})\).


\textbf{Style Disparity Loss \(\mathcal{L}_{SD}\).}
In the early phase of training, the standard deviation of \(\mathbf{w}\) is relatively low, indicating a lower confidence in the model, and its distribution closely resembles a uniform distribution\cite{jiang2023normsoftmax}. This similarity in distribution results in \(\mathbf{w}\) appearing almost indistinguishable from different watermarked fonts \(\mathbf{I}_{en^{\prime}}\), regardless of the hidden message \(\mathbf{m}\). Consequently, it becomes challenging for the decoder \(D\) to distinguish between \(\mathbf{I}_{en^{\prime}}\). This lack of discrepancy leads to a weakening of the supervision of \(D\) to \(E\), which results in unproductive training. To break this vicious circle, we propose an auxiliary loss that encourages the disparity of \(\mathbf{w}\) among different \(\mathbf{m}\) at the initial stage of training. Inspired by SupCon\cite{khosla2020supervised}, we force the disparity of \(\mathbf{w}\) to different \(\mathbf{m}\) with a contrastive learning technique, which maximizes the similarity of \(\mathbf{w}\) that have the same \(\mathbf{m}\), and minimizes the ones that are different. We index \(\mathbf{w}\) by \(b\), such that \( \mathbf{w}_{b} = \mathrm{softmax}(f_w(\bar{\mathbf{s}_t}, f_c(\mathbf{I}^{(b)}_{co}), \mathbf{m}_{b})/\tau_w) \), form the anchor index set of \( \mathbf{w}_{b} \)'s by using the other \(\mathbf{w}\) within the mini-batch \( \mathcal{A} = \mathcal{B} \setminus \{ b\} \). Also, the positive index set accommodates the \(\mathbf{w}\)'s which share the same \(\mathbf{m}\), namely, \( \mathcal{P}_1 = \{ p \in \mathcal{A} \mid \mathbf{m}_{p} = \mathbf{m}_{b}  \}  \). Then, the style disparity loss can be expressed as
\begin{equation}
    \mathcal{L}_{SD}(\mathbf{w}) = \sum_{b \in \mathcal{B}} \frac{-1}{|\mathcal{P}_1|} \sum_{p \in \mathcal{P}_1} \log \frac{\exp ( -  (\mathbf{w}_{b} - \mathbf{w}_{p})  / \tau_{sd})} {\sum_{a \in \mathcal{A}} \exp ( - (\mathbf{w}_{b} - \mathbf{w}_{a} )/ \tau_{sd} )}.
\label{eq: SD loss}
\end{equation}

\textbf{Message Reconstruction Loss \(\mathcal{L}_{MR}\).}
To recover \(\mathbf{m}\) from \(\mathbf{I}_{no}\), a CLIP training paradigm is used to maximize the similarity between font visual representations with the corresponding textual representations, while minimizing the one between unmatched pairings. We let \(\mathbf{z}_{b} = f_{ce}(\mathbf{I}^{(b)}_{no})\) and \(\mathbf{t}_k = f_{te}(Y_k)\), where \(k\) is the index of textual labels. Formally, the objective loss along the image axis can be written as
\begin{equation}
    \mathcal{L}_{I}(\mathbf{z}, \mathbf{t}) = \sum_{b \in \mathcal{B}} -\log \frac{\exp ( \mathbf{z}_b \cdot \mathbf{t}_b / \tau_{c})} {\sum_{k \in \mathcal{Y}} \exp ( \mathbf{z}_b \cdot \mathbf{t}_k / \tau_{c} )}.
\label{eq: SD loss}
\end{equation}
Similarly, we can calculate the loss along the text axis:
\begin{equation}
    \mathcal{L}_{T}(\mathbf{z}, \mathbf{t}) = \sum_{k \in \mathcal{Y}} -\log \frac{ \sum_{p \in \mathcal{P}_2} \exp ( \mathbf{t}_k \cdot \mathbf{z}_p / \tau_{c})} {\sum_{b \in \mathcal{B}} \exp ( \mathbf{t}_k \cdot \mathbf{z}_b / \tau_{c} )}, 
\label{eq: SD loss}
\end{equation}
where \(\left. \mathcal{P}_2 = \{ p \in \mathcal{B} \mid \mathbf{m}_p = \mathbf{m}_b  \}  \right.\). The entire message reconstruction loss is computed by \(\mathcal{L}_{MR} = \mathcal{L}_I + \mathcal{L}_T\).

\textbf{Style Consistent Loss \(\mathcal{L}_{SC}\).} 
We observe that the model fails to generate imperceptible variants when handling different font styles at once. By analyzing the style feature components of variants, we find that this problem is caused by the large gap between them. Inspired by \cite{xie2021dg}, we include a style consistent loss to ensure the generated style feature \(\mathbf{s}^{\prime}\) closely resembles the original feature \(\bar{\mathbf{s}}_t\), thereby ensuring imperceptible modification of fonts:
\begin{equation}
    \mathcal{L}_{SC}(\mathbf{s}^{\prime}, \bar{\mathbf{s}}_{t}) = \sum_{b \in \mathcal{B}} \|\mathbf{s}^{\prime}_b - \bar{\mathbf{s}}_{t}\|^2.
\label{eq: SC loss}
\end{equation}

\textbf{Image Reconstruction Loss \(\mathcal{L}_{IR}\).}
Perceptual loss \cite{johnson2016perceptual} is adopted to minimize the visual difference between \(\mathbf{I}_{co^{\prime}}\) and \(\mathbf{I}_{en^{\prime}}\), which is represented as
\begin{equation}
    \mathcal{L}_{IR}(\mathbf{I}_{co^{\prime}},\mathbf{I}_{en^{\prime}})= \sum_{b \in \mathcal{B}}  \left(f_\mathrm{VGG}(\mathbf{I}^{(b)}_{co^{\prime}}) - f_\mathrm{VGG}(\mathbf{I}^{(b)}_{en^{\prime}})\right)^2.
\label{perceptual loss}
\end{equation}

\textbf{Adversarial Loss \(\mathcal{L}_{G}\).}
Adversarial loss further enhances the realism of \(\mathbf{I}_{en^{\prime}}\), 
\begin{equation}
    \mathcal{L}_G(\mathbf{I}_{en^{\prime}})= \sum_{b \in \mathcal{B}} \log(1-A(\mathbf{I}^{(b)}_{en^{\prime}})).
\label{perceptual loss}
\end{equation}

Meanwhile, for the discriminator, the loss function becomes
\begin{equation}
    \mathcal{L}_A(\mathbf{I}_{co^{\prime}},\mathbf{I}_{en^{\prime}})= \sum_{b \in \mathcal{B}} \log(A(\mathbf{I}^{(b)}_{en^{\prime}})) + \log(1-A(\mathbf{I}^{(b)}_{co^{\prime}})).
\label{perceptual loss}
\end{equation}

\begin{figure*}[h]
    \centering
    \includegraphics[width=0.95\textwidth]{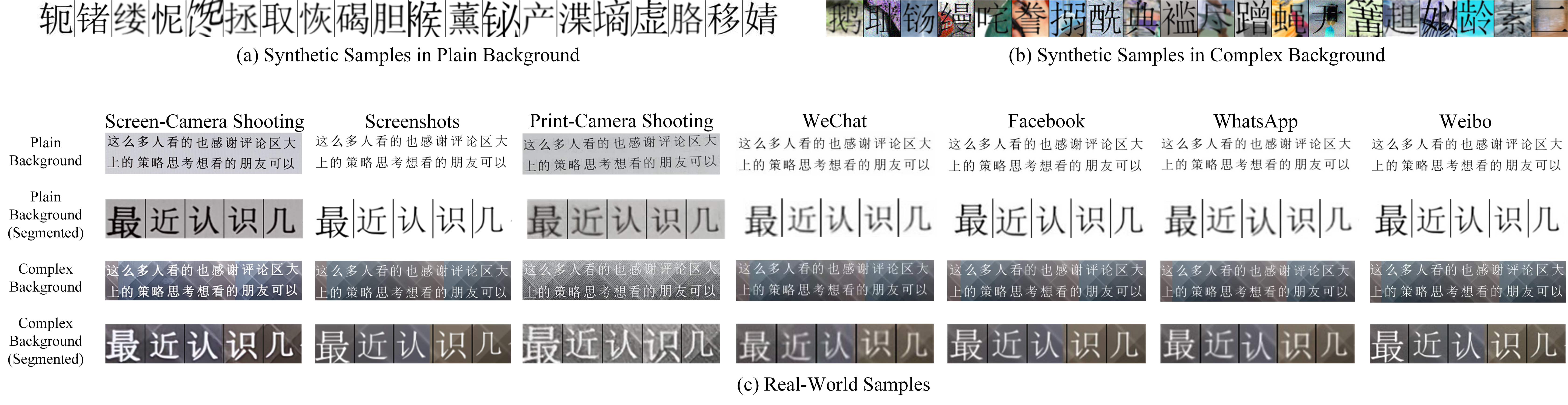}
    \caption{Synthetic samples in plain and complex backgrounds are presented in (a) and (b) respectively, while real-world samples are displayed in (c). Segmented character samples are exhibited in second and fourth rows with plain and complex backgrounds. Note that these segmented characters may not be precisely centered within the images.}
    \label{fig:test samples}
\end{figure*}

\section{Experiments}\label{sec:experiment}
In this section, we present the experimental results of FontGuard. We start by detailing the experimental setup and then present the results on the decoding accuracy and visual quality. We also explore the bit capacity and generalizability of FontGuard-GEN, together with the ablation studies.


\subsection{Experiment Settings}
\label{subsec: exp setting}

\subsubsection{Training Dataset}
We collect a font set consisting of 240 Chinese fonts and 104 English fonts\footnote{https://chinesefontdesign.com/, We exclude hieroglyphic or decorative fonts manually, as their character patterns are unpredictable.}. The Chinese vocabulary, derived from the standard character set GB/T 2312, consists of a total of 6763 characters \cite{belowcjkv}. Additionally, we include all digits and English letters to form the English vocabulary. Rendering font images to TTF files is done using FontForge\footnote{https://fontforge.org/}, an open-source font editing tool. For single-font watermarking models, we utilize character images from the specific font. To train FontGuard-GEN, we incorporate all 240 Chinese fonts. As for the background augmentation, we randomly select 5000 images from the MS COCO training set \cite{lin2014microsoft}.

\subsubsection{Test Dataset} 
We build a comprehensive test dataset to validate the robustness of font-based watermarking methods across various real-world transmission scenarios. The test set covers 7 main document distribution channels in cross-media \cite{yang2023autostegafont} and practical OSNs \cite{wu2022robust} transmissions with four different font sizes. To evaluate the watermarking methods, we embed 1000 bits into each document. For fairness in our experiments, we randomly select a paragraph from an online blog, adjusting its length to meet the requirements necessary for embedding the hidden bits. We further assess the robustness against various backgrounds by adding a complex background to the documents. After obtaining the distorted samples, CRAFT \cite{baek2019character} is used for the character segmentation, and no additional processing is applied. Character samples subjected to the seven transmission scenarios, both with a plain and complex background, are shown in Fig.~\ref{fig:test samples} (c).


\subsubsection{Implementation Details}
The base font model employed in this study is DG-Font \cite{xie2021dg}. We use all the collected fonts with all characters in the vocabulary for training the DF-Font model. All the settings follow the official code \footnote{https://github.com/ecnuycxie/DG-Font}. 
For pretraining the CLIP decoder, we adapt LASTED\footnote{https://github.com/HighwayWu/LASTED} \cite{wu2023generalizable} and use the collected font libraries for font classification. To obtain better visual results, we adopt a font image super-resolution model released by \cite{wang2021deepvecfont} to increase the resolution of the watermarked font images from size \(80 \times 80\) to \( 256 \times 256 \). The font images are then converted to SVG format. 
For the weight factors in the objectives, we empirically set \(\lambda_{SR} = 0.01\), \(\lambda_{IR} = 0.02\) and \(\lambda_{G} = 0.1\), and employ Adam \cite{kingma2014adam} as the optimizer of our models. We train FontGuard on a single NVIDIA 4090 GPU for approximately 6 hours with the SimSun font (1-bit). To ensure the reproducibility of our results, our code and dataset are available at \href{https://github.com/KAHIMWONG/FontGuard}{https://github.com/KAHIMWONG/FontGuard}.

\subsubsection{Evaluation Metrics}
We employ both pixel-level and perceptual metrics for evaluating the visual quality of watermarked font images. Pixel-level metrics, including L1, root mean square error (RMSE), and structural similarity index measure (SSIM), assess the pixel-level consistency with the original fonts. Perceptual metrics, such as FID \cite{heusel2017gans} and LPIPS \cite{zhang2018unreasonable}, evaluate the feature similarity, aligning more with human visual perception. The robustness can be reflected by the gap in bit accuracy between clean images and the images subjected to synthetic, cross-media, and OSN distortions. More specifically, the bit accuracy is formulated as:

\begin{equation}
    \text{ACC} = 1 - \frac{1}{N}\cdot HM\Big\{D(f_d( \mathbf{I}, p)), \mathbf{m}\Big\}, 
\label{eq: SD loss}
\end{equation} where \(N\) is the total number of embedded bits, \(D\) is the watermark decoder, \(f_d\) and \(p\) represent the distortion operation and its corresponding parameter, \(\mathbf{I}\) is the watermarked character image, \(\mathbf{m}\) is the ground truth bitstream, and $HM\{\cdot, \cdot\}$ returns the Hamming distance. To evaluate the decoding accuracy, \(D\) receives a clear image, making \(f_d\) an identity function. Synthetic distortions, with known \(f_d\) and \(p\), include blur, crop, JPEG, noise, random perspective, and resize. For real-world distortions, where \(f_d\) and \(p\) are unknown, we simulate complex degradations such as cross-media photographing and various OSN operations.

\subsubsection{Competitors}
We compare FontGuard with seven state-of-the-art methods, including the image-based approaches SteganoGAN\footnote{https://github.com/DAI-Lab/SteganoGAN} \cite{zhang2019steganogan} and HiNet\footnote{https://github.com/megvii-model/HINet} \cite{jing2021hinet}, as well as the font-based methods RATD \cite{qi2019robust}, LUFW \cite{yang2023language}, AutoStegaFont \cite{yang2023autostegafont}, EMCC \cite{yao2024embedding}, and ASM \cite{huang2024robust}. For RATD and EMCC, which involve manual datasets and are not publicly accessible, we align the watermarked fonts with our method and use their template-based algorithms to recover hidden bits, ensuring a fair comparison. Notably, RATD and EMCC typically require text recognition to access individual characters; following \cite{yang2023language}, we allow direct character access in these models.

\subsection{Decoding Accuracy}
\label{subsec: acc}

\begin{figure*}[ht]
    \centering
    \includegraphics[width=0.95\textwidth]{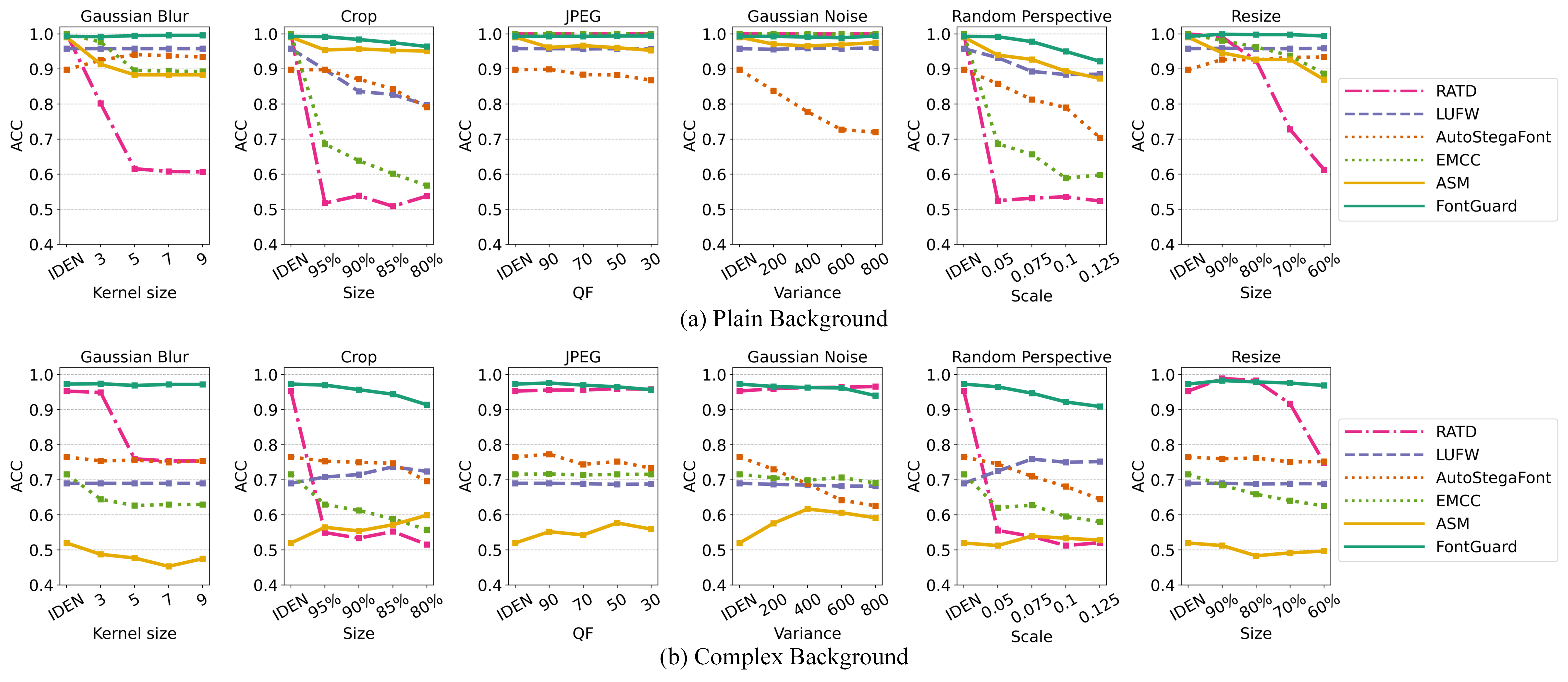}
    \caption{Decoding accuracy in (a) plain and (b) complex backgrounds across various synthetic distortions.}
    \label{fig:AUG_acc}
\end{figure*}

\setlength{\tabcolsep}{1.5pt} 
\begin{table*}[tb]
\centering
\caption{Decoding accuracy in different real-world scenarios. In each case, the highest value is in \textbf{bold}, and the second-placed is {\ul underlined}.}

\begin{tabular}{l|c|cccc|cccc|cccc|cccc|c}
    \hline
    \hline
    
    \multirow{3}{*}{Methods} & \multirow{3}{*}{Font} & \multicolumn{12}{c|}{Cross-media} & \multicolumn{4}{c|}{OSNs} & \multirow{3}{*}{Mean} \\ 

    \cline{3-18}
    
    & & \multicolumn{4}{c|}{Screen-Camera Shooting} & \multicolumn{4}{c|}{Screenshots} & \multicolumn{4}{c|}{Print-Camera Shooting} & WeChat & Facebook & WhatsApp & Weibo & \\
 
    & & 12 pt & 16 pt & 20 pt & 32 pt & 12 pt & 16 pt & 20 pt & 32 pt & 12 pt & 16 pt & 20 pt & 32 pt & 16 pt & 16 pt & 16 pt & 16 pt & \\ 

    \hline

    SteganoGAN \cite{zhang2019steganogan} & \multirow{8}{*}{SimSun} &  0.500 & 0.490 & 0.494 & \multicolumn{1}{c|}{0.496} & 0.969 & 0.979 & 0.982 & \multicolumn{1}{c|}{0.979} & 0.500 & 0.499 & 0.499 & \multicolumn{1}{c|}{0.497} & \textbf{0.978} & 0.707 & 0.761 & \multicolumn{1}{c|}{0.491} & 0.676 \\
    
    HiNet \cite{jing2021hinet} & & 0.499 & 0.499 & 0.499 & \multicolumn{1}{c|}{0.499} & \textbf{1.000} & \textbf{1.000} & \textbf{1.000} & \multicolumn{1}{c|}{\textbf{1.000}} & 0.499 & 0.499 & 0.499 & \multicolumn{1}{c|}{0.499} & 0.499 & 0.499 & 0.499 & \multicolumn{1}{c|}{0.501} & 0.624 \\

    RATD \cite{qi2019robust} & &  0.515 &  0.538 &  0.512 &  \multicolumn{1}{c|}{0.503} &  0.493 &  0.494 &  0.493 &  \multicolumn{1}{c|}{0.542} &  0.501 &  0.487 &  0.591 &  \multicolumn{1}{c|}{0.551} &  0.504 &  0.501 &  0.503 &  \multicolumn{1}{c|}{0.503} &  0.514 \\ 

    LUFW \cite{yang2023language} & & 0.641 & 0.662 & 0.672 & \multicolumn{1}{c|}{0.710} & 0.752 & 0.753 & 0.787 & \multicolumn{1}{c|}{0.813} & 0.607 & 0.623 & 0.632 & \multicolumn{1}{c|}{0.638} & 0.747 & 0.758 & 0.737 & \multicolumn{1}{c|}{0.741} & 0.705 \\

    AutoStegaFont \cite{yang2023autostegafont}& &  0.675 &  0.725 &  0.764 & \multicolumn{1}{c|}{0.752} &  0.849 &  0.874 &  0.877 & \multicolumn{1}{c|}{0.843} & 0.566 & 0.621 &  0.659 & \multicolumn{1}{c|}{0.716} &  0.764 &  0.839 & 0.597 & \multicolumn{1}{c|}{0.871} &  0.750 \\

    EMCC \cite{yao2024embedding} &  & 0.529 & 0.507 & 0.508 & \multicolumn{1}{c|}{0.521} & 0.505 & 0.508 & 0.515 & \multicolumn{1}{c|}{0.519} & 0.516 & 0.502 & 0.528 & \multicolumn{1}{c|}{0.507} & 0.490 & 0.510 & 0.485 & \multicolumn{1}{c|}{0.519} & 0.511 \\

    ASM \cite{huang2024robust} &  & {\ul0.820} & {\ul0.834} & {\ul0.858} & \multicolumn{1}{c|}{{\ul0.868}} & 0.969 & 0.952 & 0.963 & \multicolumn{1}{c|}{0.929} & {\ul0.783} & {\ul0.802} & {\ul0.842} & \multicolumn{1}{c|}{{\ul0.850}} & {\ul0.952} & {\ul0.949} & {\ul0.954} & \multicolumn{1}{c|}{{\ul0.717}} & {\ul0.878} \\

    FontGuard &  &  \textbf{0.853} &  \textbf{0.916} &  \textbf{0.951} &  \multicolumn{1}{c|}{\textbf{0.989}} &  {\ul0.987} &  {\ul0.996} &  {\ul0.991} &  \multicolumn{1}{c|}{{\ul0.992}} &  \textbf{0.872} &  \textbf{0.951} &  \textbf{0.980} &  \multicolumn{1}{c|}{\textbf{0.987}} &  0.928 &  \textbf{0.966} &  \textbf{0.972} &  \multicolumn{1}{c|}{\textbf{0.996}} &  \textbf{0.958} \\
    
    \hline

    RATD \cite{qi2019robust} & \multirow{6}{*}{FangSong} &  0.504 &  0.594 &  0.535 &  \multicolumn{1}{c|}{0.524} &  0.507 &  0.506 &  0.511 &  \multicolumn{1}{c|}{0.553} &  0.496 &  0.568 &  0.526 &  \multicolumn{1}{c|}{0.530} &  0.500 &  0.491 &  0.491 &  \multicolumn{1}{c|}{0.505} &  0.521 \\

    LUFW \cite{yang2023language} &   &  0.629 &  0.647 &  0.667 &  \multicolumn{1}{c|}{0.685} &  0.751 &  0.789 &  0.765 &  \multicolumn{1}{c|}{0.799} &  0.602 &  0.591 &  0.614 &  \multicolumn{1}{c|}{0.622} &  0.758 &  0.745 &  0.761 &  \multicolumn{1}{c|}{0.788} &  0.701 \\

    AutoStegaFont \cite{yang2023autostegafont} &   &   0.706 &   0.790 &   0.812 &  \multicolumn{1}{c|}{ 0.858} &   0.964 &   0.965 &   0.938 &  \multicolumn{1}{c|}{ 0.884} &   0.680 &   0.730 &   0.776 &  \multicolumn{1}{c|}{ 0.829} &   0.856 &   0.943 &   0.934 &  \multicolumn{1}{c|}{{\ul0.961}} &   0.852 \\

    EMCC \cite{yao2024embedding} &  & 0.504 & 0.594 & 0.535 & \multicolumn{1}{c|}{0.524} & 0.507 & 0.506 & 0.511 & \multicolumn{1}{c|}{0.553} & 0.496 & 0.568 & 0.526 & \multicolumn{1}{c|}{0.530} & 0.500 & 0.491 & 0.491 & \multicolumn{1}{c|}{0.505} & 0.521 \\
    
    ASM \cite{huang2024robust} & & {\ul0.788} & {\ul0.851} & \textbf{0.874} & \multicolumn{1}{c|}{{\ul0.939}} & {\ul0.987} & {\ul0.990} & \textbf{0.996} & \multicolumn{1}{c|}{\textbf{0.994}} & {\ul0.783} & {\ul0.867} & {\ul0.906} & \multicolumn{1}{c|}{{\ul0.962}} & \textbf{0.990} & \textbf{0.995} & \textbf{0.990} & \multicolumn{1}{c|}{0.811} & {\ul0.920} \\
        
    FontGuard & &  \textbf{0.804} &  \textbf{0.881} &  {\ul0.869} &  \multicolumn{1}{c|}{\textbf{0.962}} &  \textbf{0.988} &  \textbf{0.995} &  {\ul0.992} &  \multicolumn{1}{c|}{{\ul0.992}} &  \textbf{0.839} &  \textbf{0.921} &  \textbf{0.964} &  \multicolumn{1}{c|}{\textbf{0.975}} &  {\ul0.953} &  {\ul0.971} &  {\ul0.968} &  \multicolumn{1}{c|}{\textbf{0.992}} &  \textbf{0.942} \\

    \hline

    RATD \cite{qi2019robust} & \multirow{6}{*}{KaiTi}  &  0.506 &  0.657 &  0.484 &  \multicolumn{1}{c|}{0.515} &  0.508 &  0.509 &  0.514 &  \multicolumn{1}{c|}{0.511} &  0.504 &  0.622 &  0.514 &  \multicolumn{1}{c|}{0.515} &  0.502 &  0.507 &  0.507 &  \multicolumn{1}{c|}{0.513} &  0.524 \\

    LUFW \cite{yang2023language} &   & 0.669 & 0.646 &  0.635 &  \multicolumn{1}{c|}{0.643} &  0.733 &  0.754 &  0.737 &  \multicolumn{1}{c|}{0.738} & 0.633 &  0.625 &  0.617 &  \multicolumn{1}{c|}{0.601} &  0.739 &  0.741 &  0.733 &  \multicolumn{1}{c|}{0.759} &  0.688 \\

    AutoStegaFont \cite{yang2023autostegafont} &   &  0.616 &  0.613 &   0.701 &  \multicolumn{1}{c|}{ 0.797} &   0.836 &   0.836 &   0.795 &  \multicolumn{1}{c|}{ 0.799} &  0.615 &   0.677 &   0.694 &  \multicolumn{1}{c|}{ 0.754} &   0.808 &   0.796 &   0.852 &  \multicolumn{1}{c|}{ 0.832} &   0.751 \\

    EMCC \cite{yao2024embedding} &  & 0.482 & 0.491 & 0.536 & \multicolumn{1}{c|}{0.496} & 0.504 & 0.488 & 0.483 & \multicolumn{1}{c|}{0.524} & 0.490 & 0.493 & 0.510 & \multicolumn{1}{c|}{0.511} & 0.488 & 0.491 & 0.475 & \multicolumn{1}{c|}{0.491} & 0.497 \\

    ASM \cite{huang2024robust} &  & {\ul0.815} & {\ul0.818} & {\ul0.835} & \multicolumn{1}{c|}{{\ul0.854}} & {\ul0.990} & {\ul0.976} & {\ul0.981} & \multicolumn{1}{c|}{{\ul0.936}} & {\ul0.780} & {\ul0.789} & {\ul0.865} & \multicolumn{1}{c|}{{\ul0.864}} & {\ul0.976} & {\ul0.980} & {\ul0.977} & \multicolumn{1}{c|}{{\ul0.778}} & {\ul0.888} \\
    
    FontGuard & &  \textbf{0.877} &  \textbf{0.949} &  \textbf{0.935} &  \multicolumn{1}{c|}{\textbf{0.981}} &  \textbf{0.998} &  \textbf{1.000} &  \textbf{0.995} &  \multicolumn{1}{c|}{\textbf{0.995}} &  \textbf{0.941} &  \textbf{0.975} &  \textbf{0.988} &  \multicolumn{1}{c|}{\textbf{0.989}} &  \textbf{0.981} &  \textbf{0.997} &  \textbf{0.990} &  \multicolumn{1}{c|}{\textbf{0.999}} &  \textbf{0.974} \\

    \hline

    FontGuard & Times & 0.779 & 0.842 & 0.915 & 0.956 & 0.988 & 0.997 & 0.995 & 0.991 & 0.825 & 0.894 & 0.945 & 0.979 & 0.964 & 0.991 & 0.983 & 0.990 & 0.940 \\

    FontGuard & Verdana & 0.765 & 0.914 & 0.926 & 0.932 & 0.978 & 0.993 & 0.974 & 0.953 & 0.777 & 0.763 & 0.870 & 0.928 & 0.884 & 0.973 & 0.967 & 0.989 & 0.912 \\

    \hline
    \hline
\end{tabular}

\label{table: real acc}
\end{table*}

\underline{\textbf{Accuracy against Synthetic Distortions.}} 
To evaluate the impact of various synthetic distortions on the decoding accuracy, we subject watermarked SimSun font images to six different types of synthetic distortions. Several examples can be found in Fig.\ref{fig:test samples} (a). The bit accuracy results against synthetic distortions are shown in Fig.~\ref{fig:AUG_acc} (a). The distortion severity increases from left to right, starting from ``IDEN", which means no distortion. Additionally, we investigate the robustness against intricate backgrounds. As shown in Fig.~\ref{fig:test samples} (b), we introduce backgrounds to the font images before applying distortions. These backgrounds are randomly selected from the MS COCO validation set, with results shown in Fig.~\ref{fig:AUG_acc} (b).

We begin by evaluating the bit accuracy of each method on original font images with no distortion, namely, the cases marked with ``IDEN''. Most methods, including FontGuard, achieve a bit accuracy of over 99\%, while LUFW and AutoStegaFont attain 95.8\% and 89.8\%, respectively. Both RATD and EMCC handle pixel-level distortions, such as JPEG and noise, effectively, consistently achieving nearly 100\% bit accuracy. However, their performance drops by approximately 60\% under spatial distortions like cropping or perspective changes. This decline occurs because their reliance on pixel-level alignment for similarity scoring fails when characters are partially visible or deformed. Such a vulnerability is very serious in real-world applications, where imprecise character segmentation is not rare, adversely affecting the effectiveness of RATD and EMCC. Additionally, LUFW, AutoStegaFont, and ASM are designed for binary documents with plain backgrounds, achieving average bit accuracies of 93.0\%, 86.1\%, and 93.6\% across various distortions, respectively. However, their accuracy drops significantly on complex backgrounds, falling to 70.3\%, 72.9\%, and 53.7\%, respectively. In contrast, our FontGuard consistently outperforms other competing methods, leading to the average accuracy of 98.6\% in the plain background and 96.0\% in the complex background across various distortions. By averaging the accuracy of all the synthetic cases in the plain background, FontGuard outperforms its closest competitor, ASM, by a margin of 5.4\%, demonstrating FontGuard's superior robustness.
 
\begin{figure*}[t]
    \centering
    \includegraphics[width=0.95\textwidth]{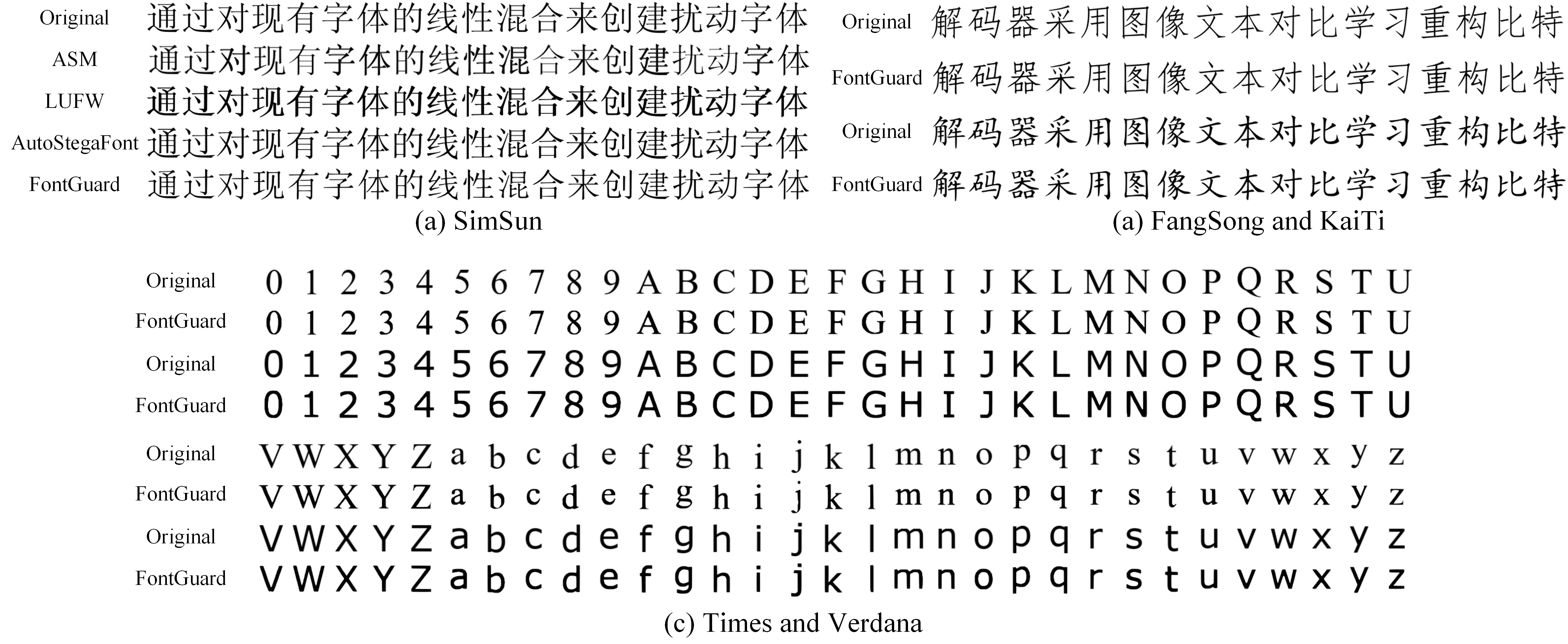}
    \caption{Examples of watermarked SimSun fonts from different methods and the original versions are depicted in (a). Examples of watermarked FangSong and KaiTi fonts are shown in (b), while watermarked Time and Verdana fonts are presented in (c). Each character is encoded with one bit, with `0' and `1' alternating in the bit sequence.}
    \label{fig:vq}
\vspace{-10pt}
\end{figure*}

\underline{\textbf{Accuracy against Real-World Distortions.}} Table~\ref{table: real acc} presents a comprehensive comparison of the decoding accuracy of watermarked fonts upon various real-world transmissions, across different fonts and font sizes. As can be seen, FontGuard consistently outperforms other methods, achieving the highest average decoding accuracy. It notably surpasses the second-best method, ASM, with considerable gains of \(+7.4\%\) and \(+5.8\%\) in cross-media and OSN scenarios respectively, while also achieving much higher visual quality as demonstrated later. In fact, in most cases, the mean accuracy of FontGuard is above 95\%. In contrast, AutoStegaFont and LUFW, with average accuracies of 78.4\% and 69.8\%, perform well in scenarios with minor background noise, such as screenshots and OSN transmissions. However, in more challenging scenarios like print-camera and screen-camera shooting, their average accuracies drop to 71.4\% and 63.8\%, respectively. Furthermore, RATD and EMCC experience a significant decline in performance compared to well-aligned cases in synthetic distortions, with accuracy around the level of a random guess at 50\%, mainly due to issues with inaccurate character segmentation. For image-based watermarking solutions like SteganoGAN and HiNet, we observe that they become inefficient when documents undergo severe distortions, such as print-camera and screen-camera shooting. Also, we conduct experiments to assess the decoding efficiency of FontGuard on various character types, including English letters and digits in Times and Verdana fonts, which are characterized by serif and sans-serif styles. As shown in the last two rows of Table~\ref{table: real acc}, FontGuard demonstrates strong performance when handling English characters, with average decoding accuracy of \(94.0\%\) and \(91.2\%\) in Times and Verdana fonts across 7 lossy channels.     

  

\subsection{Visual Quality}
\label{sec: vq}

\setlength{\tabcolsep}{1.5pt} 
\begin{table}[ht]
\centering
\caption{Quantitative comparison on the visual quality. The highest value is in \textbf{bold} and the second is {\ul underlined}.}

\begin{tabular}{l|ccccc|c}
    \hline
    \hline

    Method & L1↓ & RMSE↓ & SSIM↑ & LPIPS↓ & FID↓ & User Study(\%)↑ \\ 

    \hline
    
    CF-Font \cite{wang2023cf} & 0.060 & 0.205 & 0.754 & 0.084 & 13.130 & - \\ 
    
    \hline
    
    LUFW \cite{yang2023language} & 0.023 & 0.095 & 0.947 & 0.033 & 5.277 & {\ul 31.83} \\
    
    AutoStegaFont \cite{yang2023autostegafont} & \textbf{0.012} & \textbf{0.051} & \textbf{0.977} & \textbf{0.011} & \textbf{1.622} & \textbf{51.83} \\

    ASM \cite{huang2024robust} & 0.022  & 0.091 & 0.916 & 0.053 & 17.989 & - \\

    FontGuard & {\ul 0.019}    & {\ul 0.079}    & {\ul 0.950}    & {\ul 0.025}    & {\ul 3.606}    & \textbf{51.83}                \\ 
    
    \hline
        
    Original   & -          & -          & -          & -          & -         & 58.50 \\ 
    
    \hline
    \hline
\end{tabular}  

\label{table:vq}
\end{table}

To assess the visual quality of watermarked fonts, we conduct qualitative and quantitative comparisons between FontGuard and other methods. Fig.~\ref{fig:vq} (a) shows watermarked SimSun characters generated by four methods, alongside the original SimSun. RATD and EMCC, which involve manual font design and are not open-source, are excluded from this comparison. Fig.~\ref{fig:vq} (b) and (c) showcase examples of FontGuard applied to FangSong and KaiTi fonts, as well as Times and Verdana fonts, respectively. FontGuard consistently maintains high visual quality across all fonts. Following \cite{wang2023cf}, quantitative results of visual quality on SimSun are detailed in Table~\ref{table:vq}. FontGuard introduces slightly more perturbations than AutoStegaFont, with an increase of 0.014 in LPIPS; yet these remain subtle and typically unnoticed by users. While AutoStegaFont introduces less perturbations, it is less robust against real-world distortions, as demonstrated in Table~\ref{table: real acc}. Moreover, FontGuard significantly outperforms ASM and LUFW by 52.7\% 24.9\% in LPIPS, respectively, demonstrating its efficacy in balancing visual quality with watermark robustness.


\begin{figure}[htbp]
    \centering
    \includegraphics[width=0.9\linewidth]{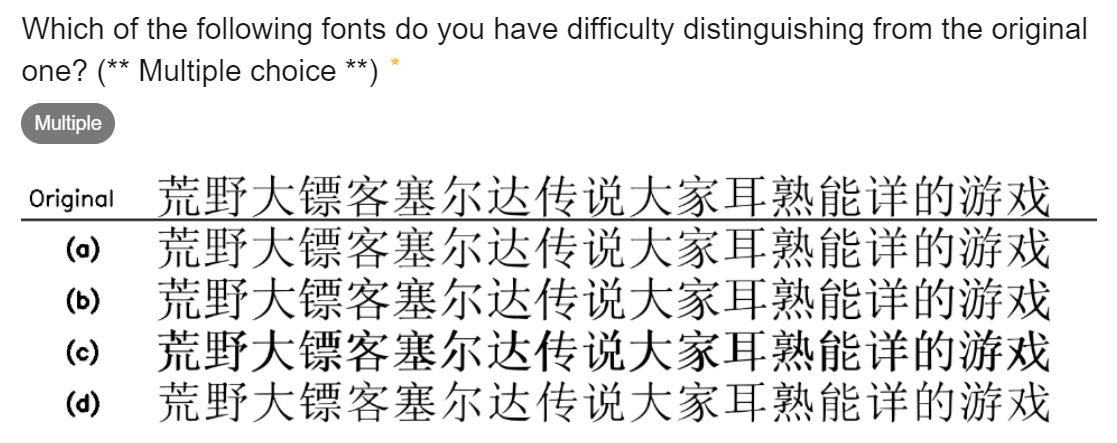}
    \caption{An example question of the user study. A reference original sentence is placed at the top of the question. Watermarked sentences of FontGuard, AutoStegaFont, and LUFW, along with the original one are shown below as choices in shuffled order.}
    \label{fig:user study}
\end{figure}

\underline{\textbf{User Study.}} We also conduct a user study to further evaluate the visual quality. We select 25 natural Chinese sentences, each containing 10-20 characters in SimSun. A sample of this study is illustrated in Fig.~\ref{fig:user study}. The study involves 24 participants who regularly use Chinese. They are asked to identify the three watermarked fonts generated by LUFW, AutoStegaFont, and FontGuard with the original. We secretly include the original font as a choice for reference. Participants are allowed multiple choices, considering they might find several fonts to be of identical quality. The study results, detailed in the last column of Table~\ref{table:vq}, show the percentage of each font being selected. The results reveal that the visual quality of FontGuard is on par with AutoStegaFont, with 51.8\% of participants selecting it, and significantly better than LUFW by a margin of 20\%. This suggests that the minor perturbations introduced by FontGuard are imperceptible to human users. Additionally, the percentage of participants choosing the original font, 58.5\%, is close to that of FontGuard, further affirming the imperceptibility of our font watermarking.

\subsection{Performance across Various Bit Capacities}
\label{sec: bit cap}

\begin{figure}[t!]
    \centering
    \includegraphics[width=\linewidth]{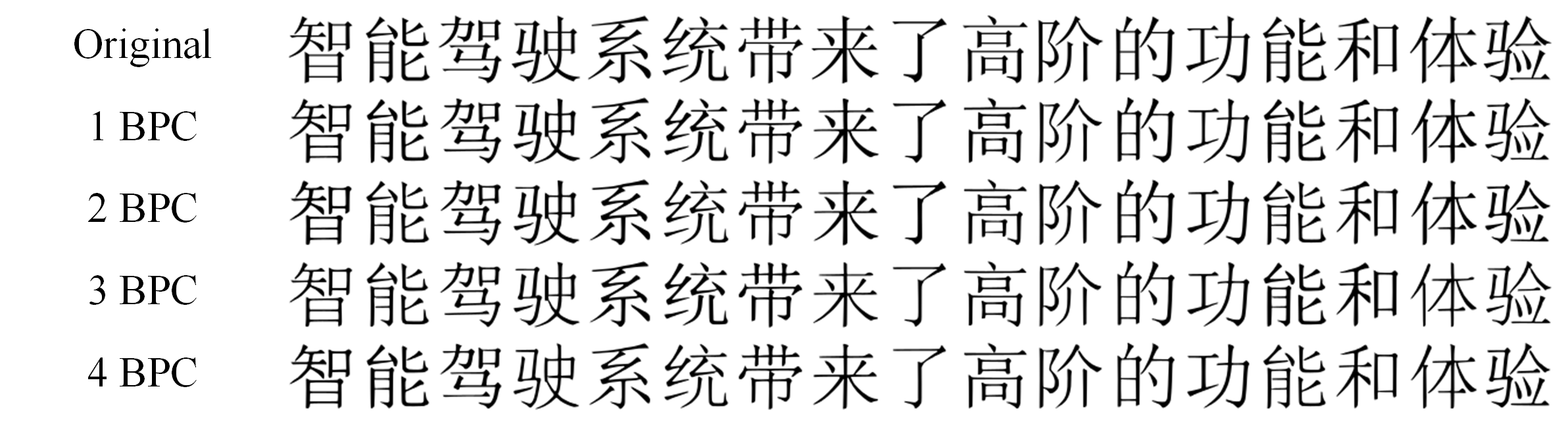}
    \vspace{-10pt}
    \caption{Examples of watermarked SimSun at different BPC. The embedded bits follow a sequence, cycling from the lowest to the highest value (in decimal) and then repeating.}
    \label{fig:diff_bit_example}
\end{figure}

\begin{figure}[t!]
    \centering
    \includegraphics[width=\linewidth]{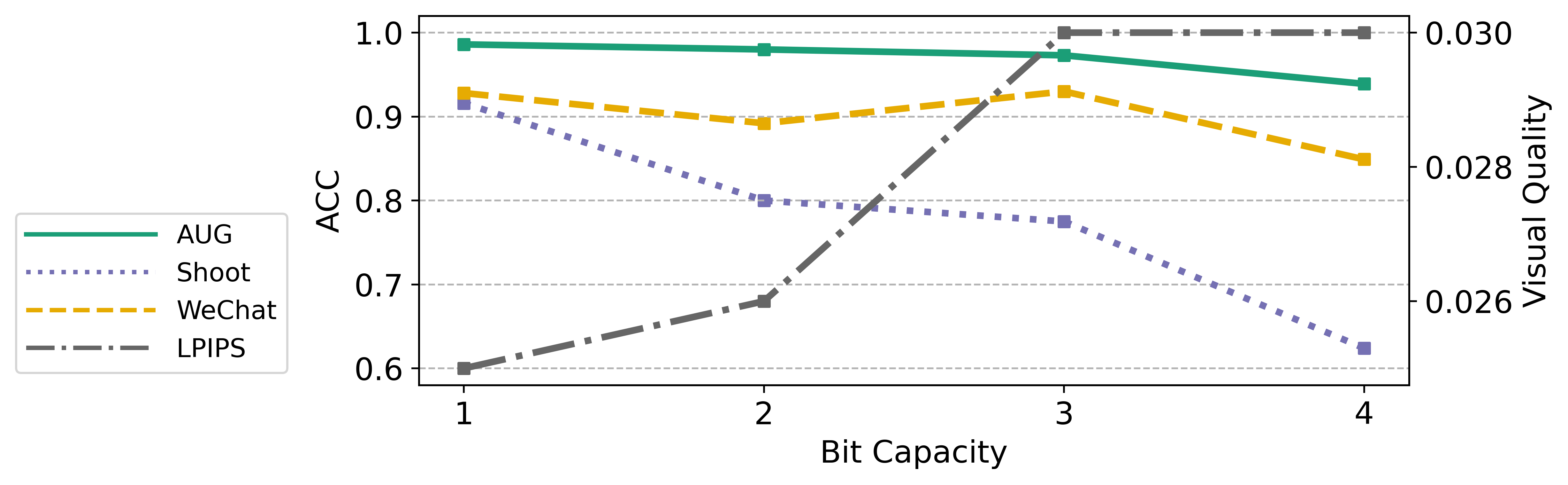}
    \vspace{-10pt}
    \caption{Overall performance across different BPC for SimSun. Three types of distortions are applied to the document: synthetic (SYN), screen-camera shooting (Shoot), and WeChat transmission. LPIPS is used as a metric to evaluate visual quality.}
    \label{fig:diff_bit}
\vspace{-10pt}
\end{figure}

Increasing BPC in font-based watermarking schemes is challenging, as the required number of variant fonts grows exponentially. Presently, LUFW and AutoStegaFont embed with 0.5 and 1 BPC, respectively. A significant advantage of FontGuard is its capacity to create a diverse range of perturbed fonts, thus enhancing its bit capacity.
Fig.~\ref{fig:diff_bit} presents the decoding accuracy with varing BPC under different scenarios. While accuracy stays around 90\% for synthetic or WeChat distortions, it becomes more susceptible to screen-camera shooting, with bit accuracy dropping from 91.6\% to 62.4\% as BPC increases from 1 to 4. This highlights a key area for improvement in maintaining high decoding accuracy and high bit capacity across various challenging scenarios.
Fig.~\ref{fig:diff_bit} also shows the impact on visual quality as BPC increases. While higher BPC introduces more perturbations, these remain comparable quality against the LUFW baseline in Table~\ref{table:vq}, with LPIPS increasing from 0.025 to 0.030 as BPC improves.
Fig.~\ref{fig:diff_bit_example} shows examples of watermarked SimSun at different BPC levels alongside the original. Remarkably, these perturbations remain imperceptible to users, even at 4 BPC, which corresponds to 16 different variants.


\subsection{Generalizability}
\label{sec: uni}

\begin{figure}[t!]
    \centering
    \includegraphics[width=0.8\linewidth]{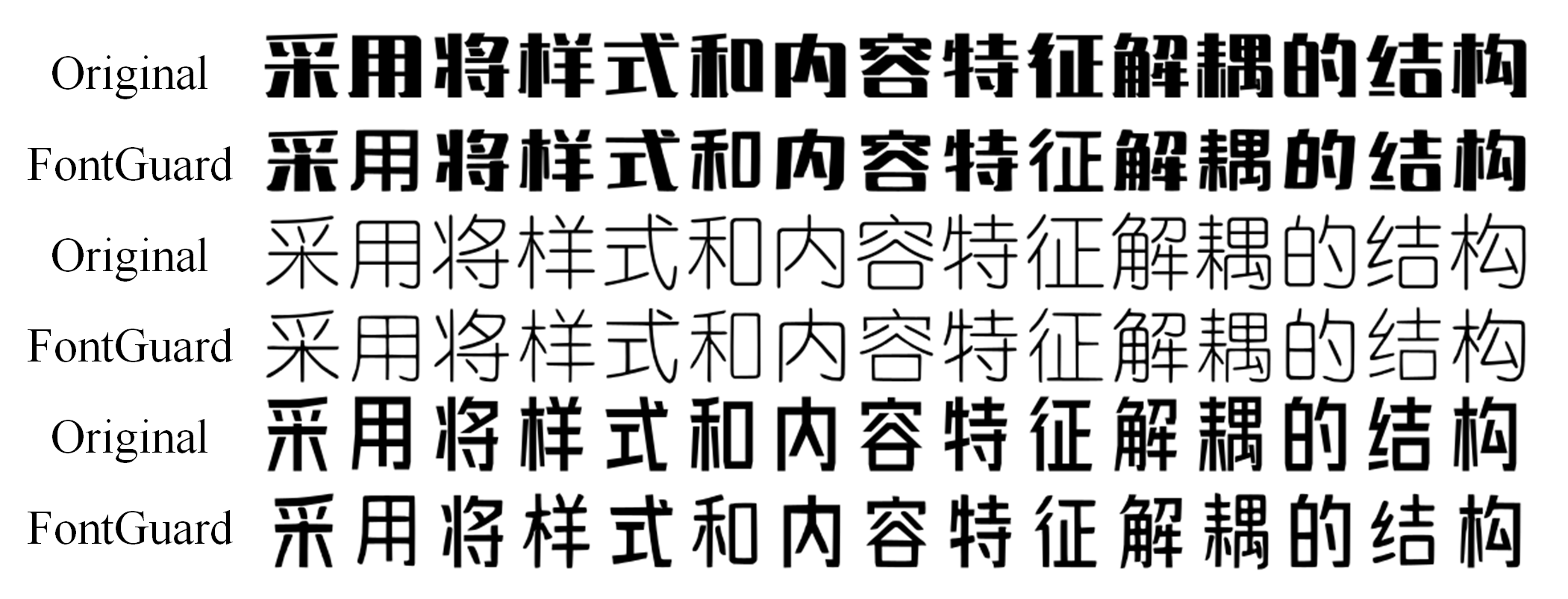}
    \caption{Examples of watermarked fonts for three unseen fonts. Each character is encoded with one bit; `0’ and `1’ are alternating.}
    \label{fig:uni_example}
\end{figure}

\setlength{\tabcolsep}{8pt} 
\begin{table}[t]
\centering
\caption{Performance across unseen fonts. RB, KSC, and CY are abbreviations of the three unseen fonts REEJIBloom, KingsoftCloud, and CloudYuanXiGBK.}

\begin{tabular}{c|cccc}
    \hline
    \hline
    
    & Font & RB & KSC & CY \\ 
    
    \hline
        
    \multirow{5}{*}{Visual Quality} & L1↓ & 0.029 & 0.032 & 0.025 \\
    
    & RMSE↓             & 0.117    & 0.123    & 0.096    \\
    
    & SSIM↑             & 0.893    & 0.895    & 0.928    \\
    
    & LPIPS↓            & 0.053    & 0.045    & 0.029    \\
    
    & FID↓              & 8.580    & 5.876    & 3.049    \\ 
    
    \hline
    
    \multirow{3}{*}{Accuracy}     & SYN               & 0.968    & 0.979    & 0.938    \\
    
    & Shoot & 0.892    & 0.839    & 0.634    \\
    
    & Wechat      & 0.950    & 0.955    & 0.864    \\ 
    
    \hline
    \hline
\end{tabular}

\label{table:uni}
\end{table}

We now evaluate FontGuard-GEN using three unseen fonts that are not included in our training set: REEJIBloom (RB), KingsoftCloud (KSC), and CloudYuanXiGBK (CY). Fig.~\ref{fig:uni_example} shows a qualitative comparison of the original and watermarked fonts, demonstrating that FontGuard-GEN effectively maintains the visual quality of each watermarked font. In addition, we give the quantitative results of the visual quality as well as the decoding accuracy in Table~\ref{table:uni}. Overall speaking, FontGuard-GEN successfully produces visually satisfactory watermarked fonts, maintaining a subtle difference from the originals compared to the LUFW baseline. A notable trade-off can be observed between the visual similarity and the decoding accuracy. Specifically, fonts with larger perturbations, such as RB and KSC (0.053 and 0.045 LPIPS) exhibits higher accuracy, achieving above 95\% accuracy in synthetic and WeChat distortions, and above 83\% in screen-camera shooting scenarios.  However, the CY font, with smaller perturbations (0.029 LPIPS), leads to inferior decoding accuracy; e.g., only 63.4\% accuracy in the camera shooting case.

\subsection{Efficiency}

\begin{table}[t]
\centering
\caption{Time overhead comparison for 1000-bits watermark embedding and detection in second.}
    \begin{tabular}{c|c|c}
        \hline
        \hline
        
        Method & Embedding (s) & Detection (s) \\
        \hline
        RATD \cite{qi2019robust}  & 1.20 & 86.27 \\
        LUFW \cite{yang2023language} & 2.40 & 4.64 \\
        AutoStegaFont \cite{yang2023autostegafont} & 1.20 & 1.56 \\
        EMCC \cite{yao2024embedding} & 1.20 & 86.73 \\
        ASM \cite{huang2024robust} & 6.09 & 0.46 \\
        FontGuard & 1.20 & 2.94 \\
    
        \hline
        \hline
    \end{tabular}
\label{table:eff}
\end{table}

In Table \ref{table:eff}, we compare the time overhead of watermark embedding and detection for our method and the competitors. Except for ASM, all other methods precompute font variants encapsulated in the TTF files, where watermark embedding is exactly the process of font variant replacement. Specifically, LUFW requires twice as many characters to embed the same amount of information as our method, resulting in \(2\times\) the time overhead. In contrast, ASM dynamically adjusts font variants based on the preceding character, consuming \(5\times\) more time than our approach. For detection, ASM stands out as the fastest due to its simple decoding process. AutoStegaFont and our method follow, leveraging GPU parallelization for efficient decoding. In comparison, LUFW and EMCC are significantly slower, as their template-based decoding process involves pixel-level comparisons of each character with multiple templates. This approach is highly time-intensive, taking over \(29\times\) longer than our method.


\subsection{Ablation Study}
\label{subsec: ablation}

\begin{figure}[t!]
    \centering
    \includegraphics[width=8.5cm]{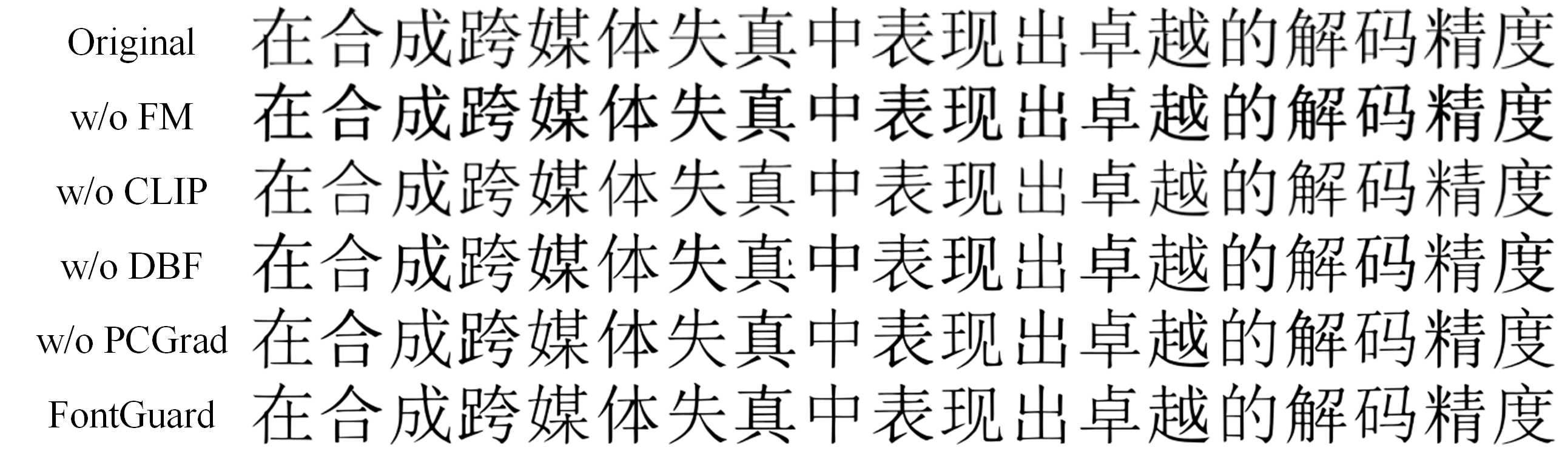}
    \caption{Examples of watermarked fonts for ablation study. Each character is encoded with one bit; `0’ and `1’ are alternating.}
    \label{fig:ablation_vq}
\vspace{-10pt}
\end{figure}

We conduct ablation studies to evaluate the impact of network structures, augmentations, and training strategies used in FontGuard to its overall performance. Fig.~\ref{fig:ablation_vq} provides examples of watermarked fonts in different model settings. 

\setlength{\tabcolsep}{6pt} 
\begin{table*}[t!]
\centering
\caption{Ablation studies in terms of the different network structures and training strategies. FM, DBF, Stage\(\mathrm{I}\), and FCP are abbreviations of font mode, differentiable binarization filter, stage\(\mathrm{I}\) training, and font classification pertaining. Note that `-' means the network is not converged.}

\begin{tabular}{cl|ccccc|ccc}
    \hline
    \hline
    
    \multirow{2}{*}{\#} & \multirow{2}{*}{Setting} & \multicolumn{5}{c}{Visual Quality} & \multicolumn{3}{c}{Accuracy} \\ 
    
    \cline{3-10}
    
    & & L1↓   & RMSE↓ & SSIM↑ & LPIPS↓ & FID↓  & SYN  & Shoot & WeChat \\ 
    
    \hline

    \#1 & w/o FM & 0.023 & 0.093 & 0.950 & 0.041 & 10.937 & 0.999 & 0.889 & 0.970 \\

    \#2 & w/o CLIP & 0.025 & 0.101 & 0.917 & 0.041 & 7.032 & 0.999 & 0.859 & 0.989 \\
    
    \#3 & w/o DBF     & 0.020 & 0.083 & 0.947 & 0.026  & 4.363 & 0.975 & 0.565 & 0.972  \\
    
    \#4 & w/o PCGrad  & 0.020 & 0.083 & 0.945 & 0.028  & 3.885 & 0.972 & 0.756 & 0.980  \\
    
    \#5 & w/o Stage \(\mathrm{I}\) & - & - & - & - & - & - & - & -  \\
    
    \#6 & w/o FCP & - & - & - & - & - & - & - & - \\
    
    \#7 & FontGuard   & 0.019 & 0.079 & 0.950 & 0.025  & 3.606 & 0.986 & 0.916 & 0.928  \\ 
    
    \hline
    \hline
\end{tabular}

\label{table:ablation}
\vspace{-10pt}
\end{table*}

\underline{\textbf{Influence of Network Structure.}}
To assess the impact of the font model in the watermarking encoder, we replace it with a CNN encoder from AutoStegaFont, resulting in a decrease in watermarked font quality with +0.016 in LPIPS, as shown in row \#1 of Table \ref{table:ablation}. This highlights the effectiveness of altering hidden style features over pixel-level alterations. 
Dropping the CLIP paradigm for the decoder, replaced with a CNN decoder in AutoStegaFont, and training with cross-entropy loss, result in degradation in both font quality and decoding robustness. Specifically, in terms of the font quality, there is an increment of 0.016 in LPIPS. Meanwhile, in challenging scenarios e.g., screen-camera shooting, the decoding accuracy experienced a \(5.7\%\) drop, as shown in the row \#2 of Table \ref{table:ablation}.
Removing the DBF also leads to a decline in the visual quality and decoding accuracy, especially in challenging situations, e.g., screen-camera shooting, where the minor font noise could disrupt decoding accuracy substantially, dropping to \(56.5\%\) as shown in row \#3.

\underline{\textbf{Influence of Training Strategy.}}
Convergence poses a significant challenge in developing our end-to-end training model. To address this, we have implemented several training strategies, and their effectiveness is evident when they are omitted, as shown in rows \#4-\#6 of Table~\ref{table:ablation}. 
Specifically, one of the employed training strategies is the PCGrad, which helps balance conflicting gradients from the divergent objectives of \( \mathcal{L}_{IR} \) and \( \mathcal{L}_{MR} \).  This balancing act, enabled by PCGrad, leads to noticeable improvements of \(+4.9\%\) in decoding accuracy on average. In addition, the two-stage training strategy is crucial for the model convergence. We analyze the style combination weights $\mathbf{w}$ and find that the distribution after the softmax has a high entropy during the initial training stage. This results in similar $\mathbf{w}$ values for different embedded bits, leading to indistinguishable images and hindering the watermarking decoder from classifying hidden bits.  Introducing the auxiliary loss $\mathcal{L}_{SD}$ encourages more distinctive $\mathbf{w}$ values for different embedded bits, and hence, the the decoder is capable of extracting features for classifying different hidden bits more effectively. Additionally, pre-training the watermarking decoder is also pivotal for the convergence. Our watermarking decoder, compared to AutoStegaFont, has more parameters in the decoder \cite{yang2023autostegafont}. This additional complexity requires extra time to converge. The feedback loop between the watermarking encoder and decoder needs better parameter initialization to prevent non-convergence issues.


\setlength{\tabcolsep}{4pt} 
\begin{table}[t!]
\centering
\caption{Ablation studies to evaluate the impact of BGA Plain and Complex, which are abbreviations of plain and complex background.}

\begin{tabular}{ll|cc|cc}
    \hline
    \hline

    \multicolumn{2}{c|}{\multirow{2}{*}{Transmission}} & \multicolumn{2}{c|}{w/ BGA} & \multicolumn{2}{c}{w/o BGA} \\
    
    \cline{3-6} 
    
    \multicolumn{2}{c|}{}                  & Plain & Complex  & Plain & Complex  \\ 
    
    \hline
    
    \multirow{3}{*}{Cross-media}   & Screen-Camera  & 0.916       & 0.842       & 0.920        & 0.853        \\
    
    & Screenshots   & 0.996  & 0.980 & 0.995  & 0.950 \\
    
    & Print-Camera & 0.951  & 0.782 & 0.914  & 0.774 \\ 
    
    \hline
    
    \multirow{4}{*}{OSNs} & WeChat         & 0.928  & 0.882 & 0.949  & 0.853 \\
    
    & Facebook       & 0.966  & 0.953 & 0.960  & 0.913 \\
    
    & WhatsApp       & 0.972  & 0.948 & 0.982  & 0.907 \\
    
    & Weibo          & 0.996  & 0.986 & 0.995  & 0.968 \\ 
    
    \hline
    
    \multicolumn{2}{c|}{Mean}              & 0.961  & 0.910 & 0.959  & 0.888 \\ 
    
    \hline
    \hline
\end{tabular}

\label{table:bga}
\vspace{-5pt}
\end{table}

\underline{\textbf{Influence of Background Augmentation.}}
Data augmentation is a proven method for enhancing the model robustness. We propose to augment the background of fonts as a novel augmentation technique specifically for the font watermarking. As shown in Table~\ref{table:bga}, BGA improves the decoding performance not only for complex background samples but also for plain background ones, yielding average gains of \(+0.2\%\) and \(+2.2\%\) across various distortions, respectively. We hypothesize that gains in plain background samples are due to the additional noise introduced during the transmission, as BGA encourages the decoder to extract background-agnostic features. This is supported by the significant improvement of \(+3.7\%\) observed in print-camera shooting scenarios.

\setlength{\tabcolsep}{1.5pt} 
\begin{table}[t!]
\centering
\caption{Ablation studies of FontGuard-GEN. Here, SCL and SP are abbreviations of Style Consistent Loss and Style Prompt.}

\begin{tabular}{ccc|ccccc|ccc}
    \hline
    \hline
    
    \multirow{2}{*}{\#} & \multicolumn{2}{c|}{Setting} & \multicolumn{5}{c|}{Visual Quality}  & \multicolumn{3}{c}{Accuracy}        \\ 
    
    \cline{2-11} 
    
    & SCL           & SP          & L1↓ & RMSE↓ & SSIM↑ & LPIPS↓ & FID↓ & SYN & Shoot & WeChat \\ 
    
    \hline
    
    \#1 &   &   & 0.071 & 0.225 & 0.740 & 0.118 & 19.938 & 0.999 & 0.958 & 1.000 \\
    
    \#2 &   & \checkmark & 0.053 & 0.185 & 0.804 & 0.087 & 10.196 & 0.997  & 0.924  & 1.000 \\
    
    \#3 & \checkmark &   & 0.028 & 0.109 & 0.911 & 0.033 & 2.641  & 0.779 & 0.752  &  0.743 \\
    
    \#4 & \checkmark & \checkmark & 0.032 & 0.123 & 0.895 & 0.045 & 5.876  & 0.979 & 0.839 & 0.955 \\ 
    
    \hline
    \hline
\end{tabular}

\label{table:ablation_uni}
\vspace{-10pt}
\end{table}

\begin{figure}[t!]
    \centering
    \includegraphics[width=0.8\linewidth]{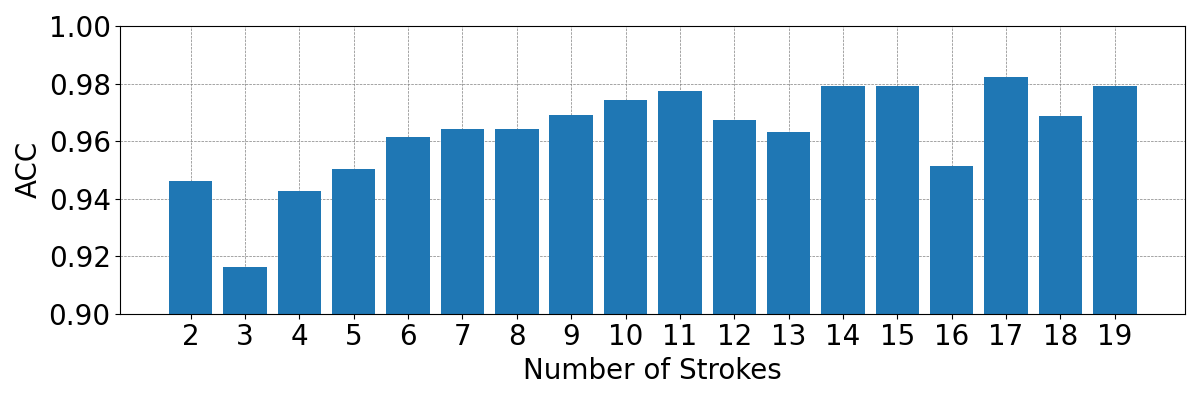}
    \caption{Accuarcy verse number of stokes in SimSun, each accuracy is computed across all real scenario cases in terms of characters with the same number of stokes.}
    \label{fig:acc_vs_stoke}
\vspace{-10pt}
\end{figure}

\underline{\textbf{Influence of Character Complexity.}}
We conduct an experiment to study the relationship between character complexity and decoding accuracy, with the results shown in Fig.~\ref{fig:acc_vs_stoke}. We observe a generally positive correlation between the number of strokes and decoding accuracy. We hypothesize that characters with more complex structures provide a greater capacity for message embedding; since the BPC is fixed, more complex characters can redundantly embed the message, thereby enhancing watermark robustness and improving accuracy.

\underline{\textbf{Influence of Settings in FontGuard-GEN.}}
To enable FontGuard to generate watermarked fonts from unseen fonts, we develop FontGuard-GEN, which incorporates the FontGuard decoder with the style prompt and the style consistent loss. Without loss of generality, We conduct ablation studies on an unseen font, KingsoftCloud, to assess the effectiveness of these components. The base version of FontGuard struggles to create high-quality fonts, as indicated by a high LPIPS of \(0.118\) (see row \#1 of Table~\ref{table:ablation_uni}). Integrating the style prompt into the decoder, as shown in row \#2, enhances the visual quality, reducing LPIPS by 0.031; but the watermarks remain noticeable. Adding the style consistent loss (row \#3) further improves the image quality and reduces LPIPS by 0.085, though the average decoding accuracy is only about 75\%. FontGuard-GEN, as outlined in row \#4, equipped with both the style prompt and style consistent loss, achieves the best tradeoff between the quality and accuracy, with the lowest LPIPS of 0.045 and a satisfactory average decoding accuracy of 92.4\%.

\section{Conclusion}\label{sec:conclusion}
In conclusion, FontGuard leverages the knowledge of a font model to produce high-quality and high-capacity watermarked fonts. Combined with a CLIP-based decoder, our FontGuard model demonstrates exceptional resilience against real-world transmission noise and achieves +5.4\%, +7.4\%, and +5.8\% accuracy improvement under synthetic, cross-media, and OSNs distortions while improving the visual
quality by 52.7\% in terms of LPIPS. The key advancements of FontGuard are the decoding robustness against severe distortions, the ability to embed more messages, and the capability to generate watermarked fonts for unseen fonts without costly re-training.




\bibliographystyle{IEEEtran}
\bibliography{ref}

\begin{thebibliography}{10}
\providecommand{\url}[1]{#1}
\csname url@samestyle\endcsname
\providecommand{\newblock}{\relax}
\providecommand{\bibinfo}[2]{#2}
\providecommand{\BIBentrySTDinterwordspacing}{\spaceskip=0pt\relax}
\providecommand{\BIBentryALTinterwordstretchfactor}{4}
\providecommand{\BIBentryALTinterwordspacing}{\spaceskip=\fontdimen2\font plus
\BIBentryALTinterwordstretchfactor\fontdimen3\font minus \fontdimen4\font\relax}
\providecommand{\BIBforeignlanguage}[2]{{%
\expandafter\ifx\csname l@#1\endcsname\relax
\typeout{** WARNING: IEEEtran.bst: No hyphenation pattern has been}%
\typeout{** loaded for the language `#1'. Using the pattern for}%
\typeout{** the default language instead.}%
\else
\language=\csname l@#1\endcsname
\fi
#2}}
\providecommand{\BIBdecl}{\relax}
\BIBdecl

\bibitem{chregular}
``Labeling of ai generated content: New guidelines released in china,'' \url{https://www.insideglobaltech.com}.

\bibitem{barrett2023identifying}
C.~Barrett, B.~Boyd, E.~Bursztein, N.~Carlini, B.~Chen, J.~Choi, A.~R. Chowdhury, M.~Christodorescu, A.~Datta, S.~Feizi \emph{et~al.}, ``Identifying and mitigating the security risks of generative ai,'' \emph{Foundations Trends Privacy Secur.}, vol.~6, no.~1, pp. 1--52, 2023.

\bibitem{usai}
``Fact sheet: Biden-⁠harris administration secures voluntary commitments from leading artificial intelligence companies to manage the risks posed by ai,'' \url{https://www.whitehouse.gov}.

\bibitem{song2025protecting}
Q.~Song, Z.~Luo, K.~C. Cheung, S.~See, and R.~Wan, ``Protecting nerfs’ copyright via plug-and-play watermarking base model,'' in \emph{Proc. Eur. Conf. Comput. Vis.}, 2025, pp. 57--73.

\bibitem{huang2025geometrysticker}
X.~Huang, K.~C. Cheung, S.~See, and R.~Wan, ``Geometrysticker: Enabling ownership claim of recolorized neural radiance fields,'' in \emph{Proc. Eur. Conf. Comput. Vis.}, 2025, pp. 438--454.

\bibitem{luo2023copyrnerf}
Z.~Luo, Q.~Guo, K.~C. Cheung, S.~See, and R.~Wan, ``Copyrnerf: Protecting the copyright of neural radiance fields,'' in \emph{Proc. IEEE Int. Conf. Comput. Vis.}, 2023, pp. 22\,401--22\,411.

\bibitem{xiao2018fontcode}
C.~Xiao, C.~Zhang, and C.~Zheng, ``Fontcode: Embedding information in text documents using glyph perturbation,'' \emph{ACM Trans. Graph.}, vol.~37, no.~2, pp. 1--16, 2018.

\bibitem{qi2019robust}
W.~Qi, W.~Guo, T.~Zhang, Y.~Liu, Z.~Guo, and X.~Fang, ``Robust authentication for paper-based text documents based on text watermarking technology,'' \emph{Math. Biosci. Eng.}, vol.~16, no.~4, pp. 2233--2249, 2019.

\bibitem{begum2020digital}
M.~Begum and M.~S. Uddin, ``Digital image watermarking techniques: a review,'' \emph{Information}, vol.~11, no.~2, p. 110, 2020.

\bibitem{yang2023language}
X.~Yang, W.~Zhang, H.~Fang, Z.~Ma, and N.~Yu, ``Language universal font watermarking with multiple cross-media robustness,'' \emph{Signal Process.}, vol. 203, p. 108791, 2023.

\bibitem{yang2023autostegafont}
X.~Yang, J.~Zhang, H.~Fang, C.~Liu, Z.~Ma, W.~Zhang, and N.~Yu, ``Autostegafont: Synthesizing vector fonts for hiding information in documents,'' in \emph{Proc. AAAI Conf. Arti. Intell.}, 2023, pp. 3198--3205.

\bibitem{huang2024robust}
X.~Huang and H.~Wang, ``Robust text watermarking based on average skeleton mass of characters against cross-media attacks,'' \emph{J. Vis. Commun. Image Representation}, vol. 104, p. 104300, 2024.

\bibitem{yao2024embedding}
Y.~Yao, C.~Wang, H.~Wang, K.~Wang, Y.~Ren, and W.~Meng, ``Embedding secret message in chinese characters via glyph perturbation and style transfer,'' \emph{IEEE Trans. Inf. Forensics and Security}, 2024.

\bibitem{sun2016processing}
W.~Sun, J.~Zhou, R.~Lyu, and S.~Zhu, ``Processing-aware privacy-preserving photo sharing over online social networks,'' in \emph{Proc. ACM Int. Conf. Multimedia}, 2016, pp. 581--585.

\bibitem{sun2018robust}
W.~Sun, J.~Zhou, S.~Zhu, and Y.~Y. Tang, ``Robust privacy-preserving image sharing over online social networks (osns),'' \emph{ACM Trans. Multim. Comput.}, vol.~14, no.~1, pp. 1--22, 2018.

\bibitem{sun2020robust}
W.~Sun, J.~Zhou, Y.~Li, M.~Cheung, and J.~She, ``Robust high-capacity watermarking over online social network shared images,'' \emph{IEEE Trans. Circuits Syst. Video Technol.}, vol.~31, no.~3, pp. 1208--1221, 2020.

\bibitem{sun2021optimal}
W.~Sun, J.~Zhou, L.~Dong, J.~Tian, and J.~Liu, ``Optimal pre-filtering for improving facebook shared images,'' \emph{IEEE Trans. Image Process.}, vol.~30, pp. 6292--6306, 2021.

\bibitem{wu2023robust}
H.~Wu, J.~Zhou, X.~Zhang, J.~Tian, and W.~Sun, ``Robust camera model identification over online social network shared images via multi-scenario learning,'' \emph{IEEE Trans. Inf. Forensics Secur.}, 2023.

\bibitem{liu2023generating}
J.~Liu, J.~Zhou, H.~Wu, W.~Sun, and J.~Tian, ``Generating robust adversarial examples against online social networks (osns),'' \emph{ACM Trans. Multim. Comput.}, vol.~20, no.~4, pp. 1--26, 2023.

\bibitem{brassil1995electronic}
J.~T. Brassil, S.~Low, N.~F. Maxemchuk, and L.~O'Gorman, ``Electronic marking and identification techniques to discourage document copying,'' \emph{IEEE J. Sel. Areas Commun.}, vol.~13, no.~8, pp. 1495--1504, 1995.

\bibitem{kim2003text}
Y.-W. Kim, K.-A. Moon, and I.-S. Oh, ``A text watermarking algorithm based on word classification and inter-word space statistics.'' in \emph{Proc. Int. Conf. Doc. Anal. Recognit.}, 2003, pp. 775--779.

\bibitem{alattar2004watermarking}
A.~M. Alattar and O.~M. Alattar, ``Watermarking electronic text documents containing justified paragraphs and irregular line spacing,'' in \emph{Secur., Steganography, Watermarking Multimedia Contents VI}, vol. 5306, 2004, pp. 685--695.

\bibitem{atallah2002natural}
M.~J. Atallah, V.~Raskin, C.~F. Hempelmann, M.~Karahan, R.~Sion, U.~Topkara, and K.~E. Triezenberg, ``Natural language watermarking and tamperproofing,'' in \emph{Int. workshop Inf. Hiding}.\hskip 1em plus 0.5em minus 0.4em\relax Springer, 2002, pp. 196--212.

\bibitem{topkara2006words}
M.~Topkara, U.~Topkara, and M.~J. Atallah, ``Words are not enough: sentence level natural language watermarking,'' in \emph{Proc. ACM Int. Workshop Contents Protection Secur.}, 2006, pp. 37--46.

\bibitem{topkara2006hiding}
U.~Topkara, M.~Topkara, and M.~J. Atallah, ``The hiding virtues of ambiguity: quantifiably resilient watermarking of natural language text through synonym substitutions,'' in \emph{Proc. workshop Multimedia Secur.}, 2006, pp. 164--174.

\bibitem{yang2020vae}
Z.-L. Yang, S.-Y. Zhang, Y.-T. Hu, Z.-W. Hu, and Y.-F. Huang, ``Vae-stega: linguistic steganography based on variational auto-encoder,'' \emph{IEEE Trans. Inf. Forensics and Security}, vol.~16, pp. 880--895, 2020.

\bibitem{yoo2023robust}
K.~Yoo, W.~Ahn, J.~Jang, and N.~Kwak, ``Robust multi-bit natural language watermarking through invariant features,'' in \emph{Proc. Annu. Meeting Assoc. Comput. Linguistics}, 2023, pp. 2092--2115.

\bibitem{peng2023text}
W.~Peng, S.~Li, Z.~Qian, and X.~Zhang, ``Text steganalysis based on hierarchical supervised learning and dual attention mechanism,'' \emph{IEEE Trans. Audio, Speech, Lang. Process.}, vol.~31, pp. 3513--3526, 2023.

\bibitem{zhu2018hidden}
J.~Zhu, R.~Kaplan, J.~Johnson, and L.~Fei-Fei, ``Hidden: Hiding data with deep networks,'' in \emph{Proc. Eur. Conf. Comput. Vis.}, 2018, pp. 657--672.

\bibitem{tancik2020stegastamp}
M.~Tancik, B.~Mildenhall, and R.~Ng, ``Stegastamp: Invisible hyperlinks in physical photographs,'' in \emph{Proc. IEEE Conf. Comput. Vis. Pattern Recogn.}, 2020, pp. 2117--2126.

\bibitem{luo2020distortion}
X.~Luo, R.~Zhan, H.~Chang, F.~Yang, and P.~Milanfar, ``Distortion agnostic deep watermarking,'' in \emph{Proc. IEEE Conf. Comput. Vis. Pattern Recogn.}, 2020, pp. 13\,548--13\,557.

\bibitem{zhong2020automated}
X.~Zhong, P.-C. Huang, S.~Mastorakis, and F.~Y. Shih, ``An automated and robust image watermarking scheme based on deep neural networks,'' \emph{IEEE Trans. Multimedia}, vol.~23, pp. 1951--1961, 2020.

\bibitem{fang2022end}
H.~Fang, Z.~Jia, Y.~Qiu, J.~Zhang, W.~Zhang, and E.-C. Chang, ``De-end: Decoder-driven watermarking network,'' \emph{arXiv preprint arXiv:2206.13032}, 2022.

\bibitem{qin2023print}
C.~Qin, X.~Li, Z.~Zhang, F.~Li, X.~Zhang, and G.~Feng, ``Print-camera resistant image watermarking with deep noise simulation and constrained learning,'' \emph{IEEE Trans. Multimedia}, vol.~26, pp. 2164--2177, 2024.

\bibitem{9956019}
H.~Fang, Z.~Jia, Y.~Qiu, J.~Zhang, W.~Zhang, and E.-C. Chang, ``De-end: Decoder-driven watermarking network,'' \emph{IEEE Trans. Multimedia}, vol.~25, pp. 7571--7581, 2023.

\bibitem{ge2023robust}
S.~Ge, Z.~Xia, J.~Fei, Y.~Tong, J.~Weng, and M.~Li, ``A robust document image watermarking scheme using deep neural network,'' \emph{Multimedia Tools Appl.}, pp. 1--24, 2023.

\bibitem{ge2023screen}
S.~Ge, J.~Fei, Z.~Xia, Y.~Tong, J.~Weng, and J.~Liu, ``A screen-shooting resilient document image watermarking scheme using deep neural network,'' \emph{IET Image Process.}, vol.~17, no.~2, pp. 323--336, 2023.

\bibitem{zhang2019steganogan}
K.~A. Zhang, A.~Cuesta-Infante, L.~Xu, and K.~Veeramachaneni, ``Steganogan: High capacity image steganography with gans,'' \emph{arXiv preprint arXiv:1901.03892}, 2019.

\bibitem{jing2021hinet}
J.~Jing, X.~Deng, M.~Xu, J.~Wang, and Z.~Guan, ``Hinet: Deep image hiding by invertible network,'' in \emph{Proc. IEEE Int. Conf. Comput. Vis.}, 2021.

\bibitem{jia2021mbrs}
Z.~Jia, H.~Fang, and W.~Zhang, ``Mbrs: Enhancing robustness of dnn-based watermarking by mini-batch of real and simulated jpeg compression,'' in \emph{Proc. ACM Int. Conf. Multimedia}, 2021, pp. 41--49.

\bibitem{wu2004data}
M.~Wu and B.~Liu, ``Data hiding in binary image for authentication and annotation,'' \emph{IEEE Trans. Multimedia}, vol.~6, no.~4, pp. 528--538, 2004.

\bibitem{cu2020robust}
V.~L. Cu, T.~Nguyen, J.-C. Burie, and J.-M. Ogier, ``A robust watermarking approach for security issue of binary documents using fully convolutional networks,'' \emph{Int. J. Doc. Anal. Recognit.}, vol.~23, pp. 219--239, 2020.

\bibitem{borges2008robust}
P.~V.~K. Borges, J.~Mayer, and E.~Izquierdo, ``Robust and transparent color modulation for text data hiding,'' \emph{IEEE Trans. Multimedia}, vol.~10, no.~8, pp. 1479--1489, 2008.

\bibitem{loc2018document}
C.~V. Loc, J.-C. Burie, and J.-M. Ogier, ``Document images watermarking for security issue using fully convolutional networks,'' in \emph{Proc. Int. Conf. Pattern Recogn.}, 2018, pp. 1091--1096.

\bibitem{cu2019hiding}
V.~L. Cu, J.-C. Burie, J.-M. Ogier, and C.-L. Liu, ``Hiding security feature into text content for securing documents using generated font,'' in \emph{Proc. Int. Conf. Doc. Anal. Recognit.}, 2019, pp. 1214--1219.

\bibitem{wu2022robust}
H.~Wu, J.~Zhou, J.~Tian, J.~Liu, and Y.~Qiao, ``Robust image forgery detection against transmission over online social networks,'' \emph{IEEE Trans. Inf. Forensics Secur.}, vol.~17, pp. 443--456, 2022.

\bibitem{radford2021learning}
A.~Radford, J.~W. Kim, C.~Hallacy, A.~Ramesh, G.~Goh, S.~Agarwal, G.~Sastry, A.~Askell, P.~Mishkin, J.~Clark \emph{et~al.}, ``Learning transferable visual models from natural language supervision,'' in \emph{Proc. Int. Conf. Mach. Learn.}, 2021, pp. 8748--8763.

\bibitem{zheng2023exif}
C.~Zheng, A.~Shrivastava, and A.~Owens, ``Exif as language: Learning cross-modal associations between images and camera metadata,'' in \emph{Proc. IEEE Conf. Comput. Vis. Pattern Recogn.}, 2023, pp. 6945--6956.

\bibitem{wu2023generalizable}
H.~Wu, J.~Zhou, and S.~Zhang, ``Generalizable synthetic image detection via language-guided contrastive learning,'' \emph{arXiv preprint arXiv:2305.13800}, 2023.

\bibitem{campbell2014learning}
N.~D. Campbell and J.~Kautz, ``Learning a manifold of fonts,'' \emph{ACM Trans. Graph.}, vol.~33, no.~4, pp. 1--11, 2014.

\bibitem{park2021few}
S.~Park, S.~Chun, J.~Cha, B.~Lee, and H.~Shim, ``Few-shot font generation with localized style representations and factorization,'' in \emph{Proc. AAAI Conf. Arti. Intell.}, 2021, pp. 2393--2402.

\bibitem{xie2021dg}
Y.~Xie, X.~Chen, L.~Sun, and Y.~Lu, ``Dg-font: Deformable generative networks for unsupervised font generation,'' in \emph{Proc. IEEE Conf. Comput. Vis. Pattern Recogn.}, 2021, pp. 5130--5140.

\bibitem{wang2023cf}
C.~Wang, M.~Zhou, T.~Ge, Y.~Jiang, H.~Bao, and W.~Xu, ``Cf-font: Content fusion for few-shot font generation,'' in \emph{Proc. IEEE Conf. Comput. Vis. Pattern Recogn.}, 2023, pp. 1858--1867.

\bibitem{yao2023vq}
M.~Yao, Y.~Zhang, X.~Lin, X.~Li, and W.~Zuo, ``Vq-font: Few-shot font generation with structure-aware enhancement and quantization,'' \emph{arXiv preprint arXiv:2308.14018}, 2023.

\bibitem{10356848}
M.~Zhao, X.~Qi, Z.~Hu, L.~Li, Y.~Zhang, Z.~Huang, and X.~Yu, ``Calligraphy font generation via explicitly modeling location-aware glyph component deformations,'' \emph{IEEE Trans. Multimedia}, pp. 1--13, 2023.

\bibitem{carlier2020deepsvg}
A.~Carlier, M.~Danelljan, A.~Alahi, and R.~Timofte, ``Deepsvg: A hierarchical generative network for vector graphics animation,'' pp. 16\,351--16\,361, 2020.

\bibitem{wang2021deepvecfont}
Y.~Wang and Z.~Lian, ``Deepvecfont: Synthesizing high-quality vector fonts via dual-modality learning,'' \emph{ACM Trans. Graph.}, vol.~40, no.~6, pp. 1--15, 2021.

\bibitem{wang2023deepvecfont}
Y.~Wang, Y.~Wang, L.~Yu, Y.~Zhu, and Z.~Lian, ``Deepvecfont-v2: Exploiting transformers to synthesize vector fonts with higher quality,'' in \emph{Proc. IEEE Conf. Comput. Vis. Pattern Recogn.}, 2023, pp. 18\,320--18\,328.

\bibitem{yu2020gradient}
T.~Yu, S.~Kumar, A.~Gupta, S.~Levine, K.~Hausman, and C.~Finn, ``Gradient surgery for multi-task learning,'' \emph{Proc. Adv. Neural Inf. Process. Syst.}, pp. 5824--5836, 2020.

\bibitem{jiang2023normsoftmax}
Z.~Jiang, J.~Gu, and D.~Z. Pan, ``Normsoftmax: Normalizing the input of softmax to accelerate and stabilize training,'' in \emph{IEEE Int. Conf. Omni-layer Intell. Syst.}, 2023, pp. 1--6.

\bibitem{khosla2020supervised}
P.~Khosla, P.~Teterwak, C.~Wang, A.~Sarna, Y.~Tian, P.~Isola, A.~Maschinot, C.~Liu, and D.~Krishnan, ``Supervised contrastive learning,'' \emph{Proc. Adv. Neural Inf. Process. Syst.}, pp. 18\,661--18\,673, 2020.

\bibitem{johnson2016perceptual}
J.~Johnson, A.~Alahi, and L.~Fei-Fei, ``Perceptual losses for real-time style transfer and super-resolution,'' in \emph{Proc. Eur. Conf. Comput. Vis.}, 2016, pp. 694--711.

\bibitem{belowcjkv}
V.~C. A.~L. Below, ``Cjkv information processing chinese japanese korean and vietnamese computing.''

\bibitem{lin2014microsoft}
T.-Y. Lin, M.~Maire, S.~Belongie, J.~Hays, P.~Perona, D.~Ramanan, P.~Doll{\'a}r, and C.~L. Zitnick, ``Microsoft coco: Common objects in context,'' in \emph{Proc. Eur. Conf. Comput. Vis.}, 2014, pp. 740--755.

\bibitem{baek2019character}
Y.~Baek, B.~Lee, D.~Han, S.~Yun, and H.~Lee, ``Character region awareness for text detection,'' in \emph{Proc. IEEE Conf. Comput. Vis. Pattern Recogn.}, 2019, pp. 9365--9374.

\bibitem{kingma2014adam}
D.~P. Kingma and J.~Ba, ``Adam: a method for stochastic optimization,'' \emph{arXiv preprint arXiv:1412.6980}, 2014.

\bibitem{heusel2017gans}
M.~Heusel, H.~Ramsauer, T.~Unterthiner, B.~Nessler, and S.~Hochreiter, ``Gans trained by a two time-scale update rule converge to a local nash equilibrium,'' \emph{Proc. Adv. Neural Inf. Process. Syst.}, vol.~30, 2017.

\bibitem{zhang2018unreasonable}
R.~Zhang, P.~Isola, A.~A. Efros, E.~Shechtman, and O.~Wang, ``The unreasonable effectiveness of deep features as a perceptual metric,'' in \emph{Proc. IEEE Conf. Comput. Vis. Pattern Recogn.}, 2018, pp. 586--595.

\end{thebibliography}

\newpage

 




\vspace{11pt}

\begin{IEEEbiography}[{\includegraphics[width=1in,height=1.25in,clip,keepaspectratio]{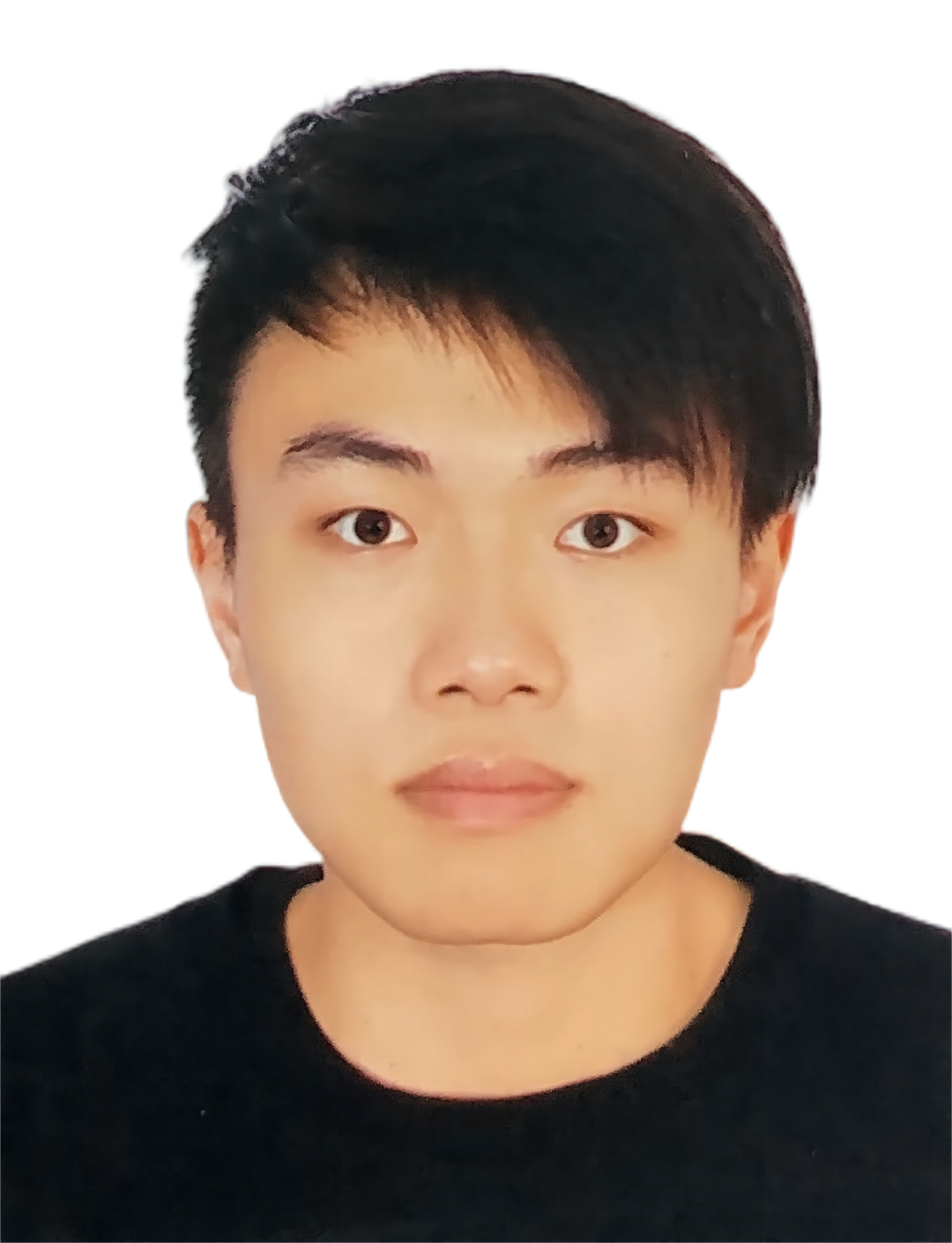}}]{Kahim Wong} received the B.S. degree in Computer Science and Technology from Wuhan University, in 2022, and the M.S. degree in Data Science from the University of Bristol in 2022. He is currently pursuing the Ph.D. degree in Computer Science at the University of Macau. His research interests include Text Watermarking and Image Forgery Detection.
\end{IEEEbiography}

\begin{IEEEbiography}[{\includegraphics[width=1in,height=1.25in,clip,keepaspectratio]{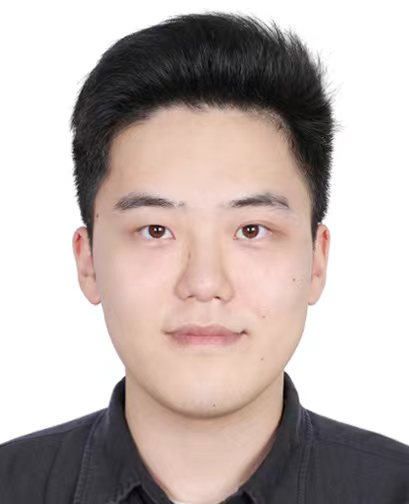}}]{Jicheng Zhou}
received the B.S. degree in Computer Science from the University of Macau in 2023. He is currently pursuing the M.S. degree at the University of Macau. His research interests include AI Security and Forensics.
\end{IEEEbiography}

\begin{IEEEbiography}[{\includegraphics[width=1in,height=1.25in,clip,keepaspectratio]{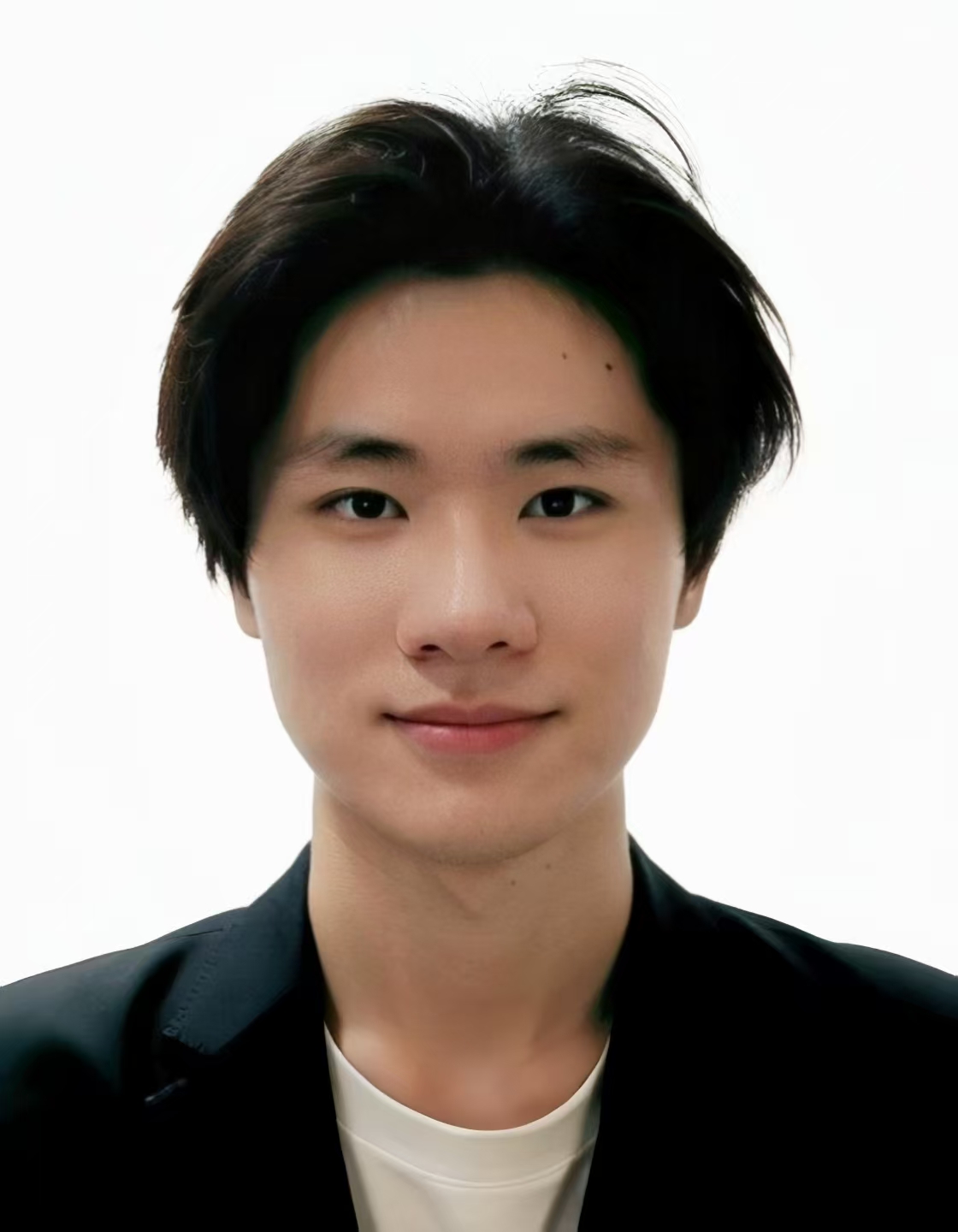}}]{Kemou Li} received the B.S. degree in mathematics and applied mathematics from Sun Yat-sen University, in 2021, and the M.S. degree in artificial intelligence applications from University of Macau, in 2023. He is currently pursuing the Ph.D. degree in computer science at University of Macau. His research interests include trustworthy machine learning and AI security.
\end{IEEEbiography}

\begin{IEEEbiography}[{\includegraphics[width=1in,height=1.25in,clip,keepaspectratio]{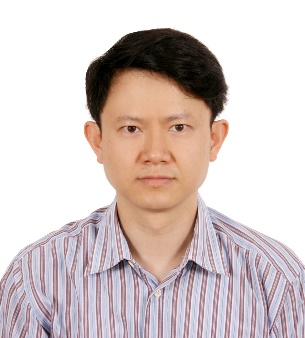}}]{Yain-Whar Si} is an Associate Professor at the Department of Computer and Information Science, University of Macau, China. His research interests include Data Analytics, Computational Intelligence, Financial Technology, and Information Visualization. He received his PhD degree from Queensland University of Technology, Australia.
\end{IEEEbiography}

\begin{IEEEbiography}[{\includegraphics[width=1in,height=1.25in,clip,keepaspectratio]{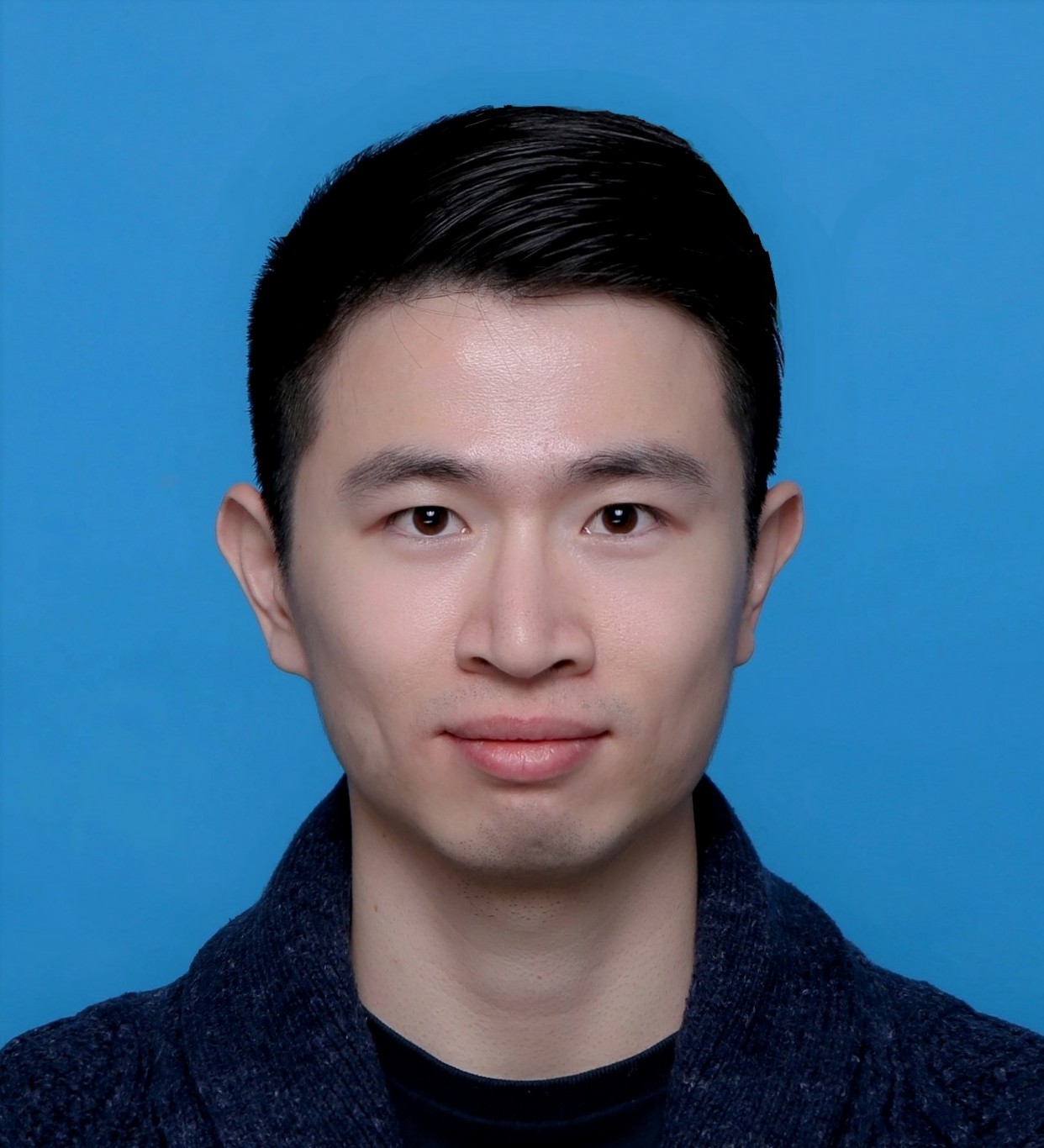}}]{Xiaowei Wu} is an Assistant Professor in the Department of Computer and Information Science with the State Key Laboratory of Internet of Things for Smart City (IOTSC) at the University of Macau. He received his Ph.D. degree from the University of Hong Kong (HKU) and his B.Eng. degree from the University of Science and Technology of China (USTC). His research interests span various topics in Multimedia Processing, Approximation Algorithms, and Algorithmic Game Theory.
\end{IEEEbiography}

\begin{IEEEbiography}[{\includegraphics[width=1in,height=1.25in,clip,keepaspectratio]{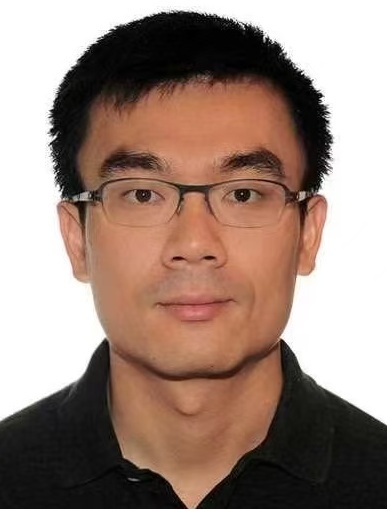}}]{Jiantao Zhou}
received the B.Eng. degree from the Department of Electronic Engineering, Dalian University of Technology, in 2002, the M.Phil. degree from the Department of Radio Engineering, Southeast University, in 2005, and the Ph.D. degree from the Department of Electronic and Computer Engineering, Hong Kong University of Science and Technology, in 2009. He held various research positions with the University of Illinois at Urbana-Champaign, Hong Kong University of Science and Technology, and McMaster University. He is now a Professor and Head of the Department of Computer and Information Science, Faculty of Science and Technology, University of Macau. His research interests include multimedia security and forensics, multimedia signal processing, artificial intelligence and big data. He holds four granted U.S. patents and two granted Chinese patents. He has co-authored two papers that received the Best Paper Award at the IEEE Pacific-Rim Conference on Multimedia in 2007 and the Best Student Paper Award at the IEEE International Conference on Multimedia and Expo in 2016. He is serving as an Associate Editor for the IEEE TRANSACTIONS ON MULTIMEDIA and the IEEE TRANSACTIONS ON DEPENDABLE and SECURE COMPUTING. 
\end{IEEEbiography}

\vfill

\end{document}